\documentclass{article}

\PassOptionsToPackage{sort,compress}{natbib}
\usepackage{natbib}

\newif\ifarxiv
\arxivtrue

\ifarxiv
\usepackage{libertine}
\usepackage[margin=1in]{geometry}
\else
\usepackage[final]{neurips_2022}
\fi

\usepackage[utf8]{inputenc} %
\usepackage[T1]{fontenc}    %
\usepackage{url}            %
\usepackage{booktabs}       %
\usepackage{amsfonts}       %
\usepackage{amsmath}
\usepackage{nicefrac}       %
\usepackage{microtype}      %
\usepackage{xcolor}         %
\usepackage{macros}
\usepackage{subcaption}
\usepackage{booktabs}
\usepackage{caption}
\usepackage{amsthm}
\usepackage{float}
\usepackage{wrapfig}

\definecolor{mydarkblue}{rgb}{0,0.08,0.45}
\usepackage[hidelinks,
colorlinks=true,
linkcolor=mydarkblue,
citecolor=mydarkblue,
filecolor=mydarkblue,
urlcolor=mydarkblue,
backref=section]{hyperref}       %
\usepackage{cleveref}

\title{Benign, Tempered, or Catastrophic:\\ A Taxonomy of Overfitting}

\author{%
  \textbf{Neil Mallinar}\thanks{Co-first authors.} \\
  UC San Diego \\
  \texttt{\sf nmallina@ucsd.edu} \\
  \and 
  \textbf{James B. Simon}\footnotemark[1] \\
  UC Berkeley \\
  \texttt{\sf james.simon@berkeley.edu} \\
  \and 
  \textbf{Amirhesam Abedsoltan} \\
  UC San Diego \\
  \texttt{\sf aabedsoltan@ucsd.edu} \\
  \and 
  \textbf{Parthe Pandit} \\
  UC San Diego \\
  \texttt{\sf parthepandit@ucsd.edu} \\
  \and 
  \textbf{Mikhail Belkin} \\
  UC San Diego \\
  \texttt{\sf mbelkin@ucsd.edu} \\
  \and 
  \textbf{Preetum Nakkiran} \\
  Apple \& UC San Diego \\
  \texttt{\sf preetum@apple.com} \\
}

\date{}

\begin{document}

\maketitle

\begin{abstract}

The practical success of overparameterized neural networks has motivated the recent scientific study of \textit{interpolating methods}, which perfectly fit their training data.
Certain interpolating methods, including neural networks, can fit noisy training data without catastrophically bad test performance, in defiance of standard intuitions from statistical learning theory.
Aiming to explain this, a body of recent work has studied \textit{benign overfitting}, a phenomenon where some interpolating methods approach Bayes optimality, even in the presence of noise.
In this work we argue that while benign overfitting has been instructive and fruitful to study, many real interpolating methods like neural networks \textit{do not fit benignly}: modest noise in the training set causes nonzero (but non-infinite) excess risk at test time, implying these models are neither benign nor catastrophic but rather fall in an intermediate regime.
We call this intermediate regime \textit{tempered overfitting}, and we initiate its systematic study.
We first explore this phenomenon in the context of kernel (ridge) regression (KR) by obtaining conditions on the ridge parameter and kernel eigenspectrum under which KR exhibits each of the three behaviors.
We find that kernels with powerlaw spectra, including Laplace kernels and ReLU neural tangent kernels, exhibit tempered overfitting.
We then empirically study deep neural networks through the lens of our taxonomy, and find that those trained to interpolation are tempered, while those stopped early are benign.
We hope our work leads to a more refined understanding of overfitting in modern learning.

\end{abstract}

\section{Introduction}

In the last decade, the dramatic success of overparameterized deep neural networks (DNNs) has inspired the field to reexamine the theoretical foundations of generalization.
Classical statistical learning theory suggests that an algorithm which \emph{interpolates} (i.e. perfectly fits) its training data will typically \emph{catastrophically overfit} at test time, generalizing no better than a random
\ifarxiv
function\footnote{
There are various ways to formalize this prediction depending on the setting:
it is a consequence of the ``bias-variance tradeoff'' in statistics, the ``bias-complexity tradeoff'' in PAC learning,
and ``capacity control''-based generalization bounds in kernel ridge regression.
.
}
\else
function.
\fi

\begin{figure}[t]
    \centering
    \includegraphics[width=\textwidth]{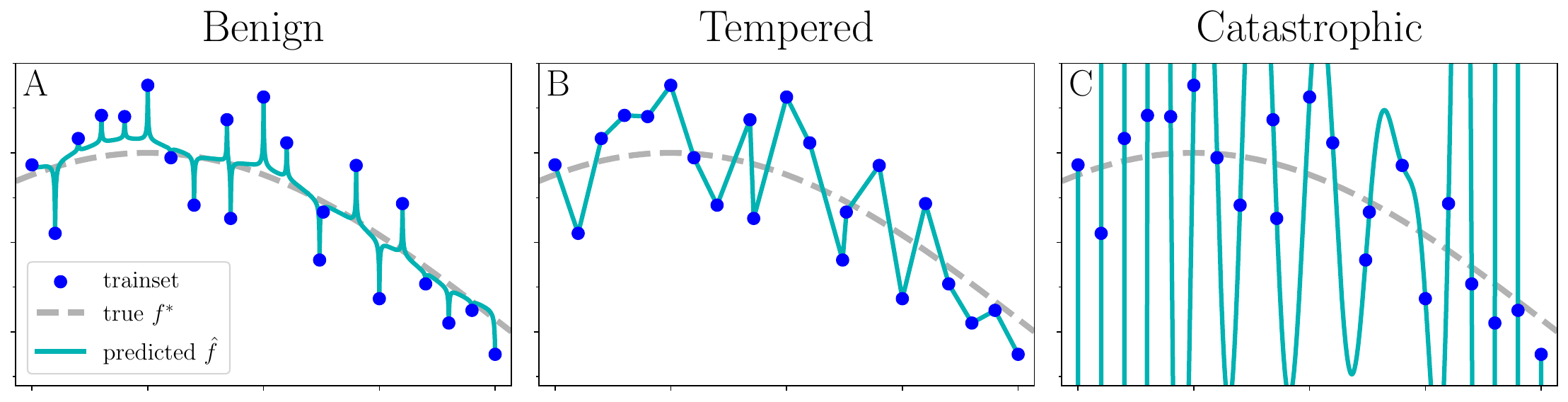}
    \caption{
    \textbf{As $n \rightarrow \infty$, interpolating methods can exhibit three types of overfitting.}
    \textbf{(A)} In \textit{benign overfitting}, the predictor asymptotically approaches the ground-truth, Bayes-optimal function.
    Nadaraya-Watson kernel smoothing with a singular kernel, shown here, is asymptotically benign.
    \textbf{(B)} In \textit{tempered overfitting}, the regime studied in this work, the predictor approaches a constant test risk greater than the Bayes-optimal risk.
    Piecewise-linear interpolation is asymptotically tempered.
    \textbf{(C)} In \textit{catastrophic overfitting}, the predictor generalizes arbitrarily poorly.
    Rank-$n$ polynomial interpolation is asymptotically catastrophic.
    }
    \label{fig:trichotomy_toy_figure}
\end{figure}

Figure \ref{fig:trichotomy_toy_figure}c illustrates the catastrophic overfitting classically expected of an interpolating method.
Defying this picture, DNNs can interpolate their training data and generalize well nonetheless \citep{DBLP:journals/corr/NeyshaburTS14,zhang2016understanding}, suggesting the need for a new theoretical paradigm within which to understand their overfitting.

This need motivated the identification and study of \textit{benign overfitting} using the terminology of~\citep{bartlett2019benign}
(also called ``harmless interpolation'' \citep{muthukumar2020harmless}), a phenomenon in which certain methods that perfectly fit the training data still approach \textit{Bayes-optimal} generalization in the limit of large trainset size.
Intuitively speaking, benignly-overfitting methods fit the target function globally, yet fit the noise only locally, and the addition of more label noise does not asymptotically degrade generalization.
Figure \ref{fig:trichotomy_toy_figure}a illustrates a simple method that is asymptotically benign\footnote{As a first hint that important practical methods may not be benign (at least in low dimension), note that, in order to be benign, the predicted function in Figure \ref{fig:trichotomy_toy_figure}a \textit{has} to take this spiky shape. On the other hand, very wide and deep neural network may indeed be spiky~\cite{radha:2022-wide-and-deep}.}.
The study of benign overfitting has proven fruitful, leading to
rich mathematical insights into high-dimensional learning\footnote{A partial list of works here include
\citet{advani2017high,d2020triple,bahri2021explaining,bahri2020statistical,goldt2019dynamics,bartlett2019benign,bartlett2021deep,belkin2018reconciling,belkin2018overfitting,belkin2018understand,cao2022benign,chatterji2021foolish,chatterji2021interplay,frei2022benign,hastie2019surprises,koehler2021uniform,liang2018just,liang2020just,mei2019generalization,muthukumar2020harmless,rakhlin2018consistency,tsigler2020benign,zhang2016understanding,zhang2021understanding}.},
and benign overfitting is certainly closer to the real behavior of DNNs than catastrophic overfitting.

\begin{wrapfigure}{r}{0.44\textwidth}
\vspace{-2em}
    \centering
    \includegraphics[width=0.44\textwidth]{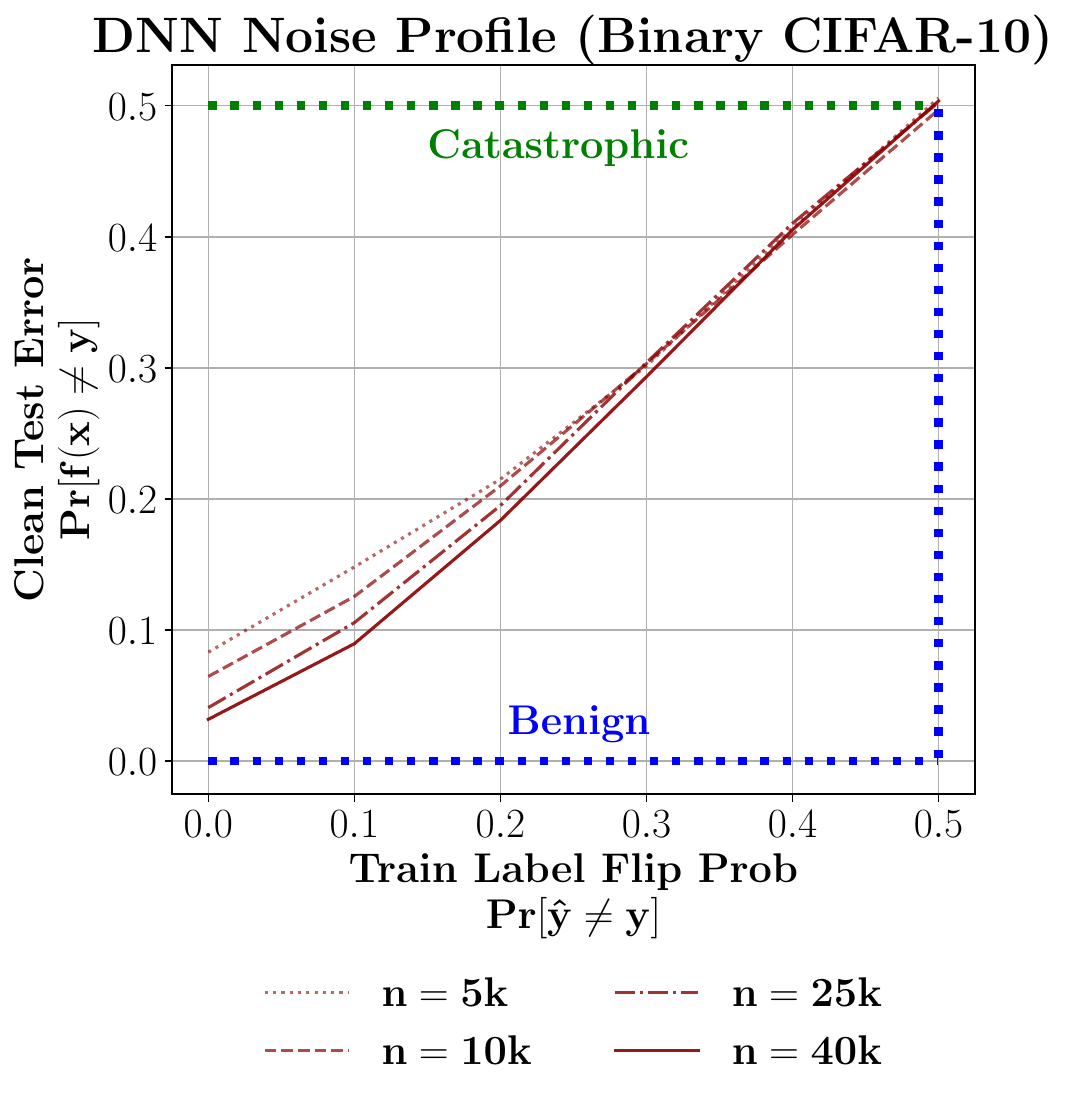}
    \caption{
        \textbf{DNNs trained on image data exhibit \textit{tempered overfitting}, not benign overfitting.}
        Curves show test classification error vs. training label flip probability for a Wide ResNet $28 \times 10$ trained to interpolation on binary CIFAR-10 (animals vs. vehicles) for different training sizes $n$.
        These curves are \textit{noise profiles}, as discussed in Section \ref{sec:prelims}.}
    \label{fig:intro_profile}
\end{wrapfigure}

That said, it requires only simple experiments to reveal that many standard DNNs \textit{do not overfit benignly}: when training on noisy data, DNNs do not diverge catastrophically, but \textit{neither} do they approach Bayes-optimal risk.
Instead, they converge to a predictor that is neither catastrophic nor optimal but rather somewhere in between, with error that increases as the noise in the data increases.
Figure \ref{fig:intro_profile} depicts such an experiment: a ResNet is trained on a binary variant of CIFAR-10 with varying amounts of training label noise, and with increasing sample size $n$.
We see from Figure~\ref{fig:intro_profile} that greater train noise indeed results in greater test error,
and this test error persists even as $n$ grows, converging to a non-zero asymptotic value\footnote{
It is well-known and is perhaps unsurprising that interpolating DNNs are harmed by label noise (e.g. \citet{zhang2016understanding}); our new observation is that this persists \textit{even as $n \rightarrow \infty$}.
}.
This is unlike ``benign overfitting,'' which would produce an asymptotically-optimal predictor at all non-trivial noise levels (depicted in blue in Figure~\ref{fig:intro_profile}).
This suggests that, in the search for a paradigm to understand modern interpolating methods, we should identify and study a regime intermediate between benign and catastrophic.

\subsection{Summary of Contributions}
In this work we formally identify an intermediate regime between benign and catastrophic overfitting.
We call this intermediate behavior \textit{tempered overfitting} because the noise's harmful effect is tempered but still nonzero.
We find that both DNNs trained to interpolation and (ridgeless) kernel regression (KR) using certain common kernels fall into this intermediate regime \textit{even as the number of training examples $n$ approaches infinity}, as do common methods like 1-nearest-neighbors and piecewise-linear interpolation (as in Figure \ref{fig:trichotomy_toy_figure}b).
Our tempered regime completes the taxonomy of overfitting:
essentially any learning procedure is either benign, tempered, or catastrophic in the asymptotic limit.

We begin in Section \ref{sec:prelims} with preliminaries, formal definitions of the three regimes, and a taxonomy of some common ML methods according to these regimes.
\ifarxiv
We generally consider an algorithm's limiting behavior as $n \rightarrow \infty$, as in the study of statistical consistency\footnote{We note that benign overfitting is statistical consistency with the  additional requirement of interpolation.}.
We are more interested in algorithms' asymptotic test risk than in the fact that they interpolate, and we consider certain non-interpolating methods as well as interpolating methods.
\fi
In Section \ref{sec:kernel}, we study these three regimes for kernel regression (KR).
Using recent spectral theories characterizing the expected test error of KR, we obtain conditions on the ridge parameter and kernel eigenspectrum under which KR falls into each of the three regimes.
Importantly, we find that ridgeless kernels with powerlaw spectra, including the Laplace kernel and ReLU fully-connected neural tangent kernels (NTKs), are asymptotically \textit{tempered}, not benign.
We confirm our theory with experiments on synthetic data.
In Section \ref{sec:experiments}, we empirically study overfitting for DNNs. 
We give evidence that standard DNNs trained to interpolation exhibit tempered overfitting, not benign overfitting, motivating the further study of tempered overfitting in the pursuit of understanding modern machine learning methods.
We additionally study the time dynamics of overfitting, and the effect of early-stopping.
We conclude with discussion in Sections \ref{sec:limitations} and \ref{sec:conclusions}.

\subsection{Relation to Prior Work}

Our work is inspired by recent developments in the theory and empirics of 
interpolating methods \citep{belkin:2019-interp-and-optimality,belkin2018does,belkin2018overfitting,bartlett2019benign,rakhlin2018consistency,ji2021early,devroye1998hilbert,koehler2021uniform,liang2018just,liang2020just,muthukumar2020harmless,tsigler2020benign}.
Many of these works prove that certain interpolating estimators
are statistically consistent in certain settings, demonstrating
benign overfitting.
In contrast, we argue that a wide range of interpolators,
including DNNs used in practice, are \emph{not benign}.
Our work is compatible with prior work because we consider different settings---erring more on the side of realism---which leads us to different conclusions.
The empirical observation that interpolating DNNs for classification are inconsistent,
and thus not benign, was made in \citet{nakkiran2020distributional}, which in part inspired the present work.

Briefly, many prior works assume that the target task always lies in ``high enough dimension''
relative to the number of samples $n$, while our work considers the limit $n \to \infty$
for tasks of fixed dimension $d$, the realistic asymptotic in practice.
For example, in the linear regression setting, \citet{bartlett2019benign} highlight that benign overfitting occurs most robustly when \emph{input dimension grows faster than the sample size}.
Other works explicitly scale the ambient problem dimension and sample size to infinity together at a proportional rate \citep{hastie2019surprises,mei2019generalization,liang2018just}.
This joint scaling is also often considered in statistical physics approaches to learning dynamics
(see \citet{zdeborova2016statistical} and references therein).
Taking a different approach, \citet{frei2022benign} prove that interpolating two-layer networks can achieve close to optimal test error on certain distributions, but require an assumption that $n \leq \Omega(d)$.
And \citet{koehler2021uniform} state a generalization bound for interpolators that decays to $0$ with $n$, but also requires $n \leq d$.
In summary, a bulk of prior benign overfitting results apply in the regime where $n$ is large,
but still restricted to be smaller than the dimension of the problem,
whereas we do not consider such restrictions.

Our work is also compatible with prior works which observe that excess risk of certain interpolating methods
decays with ambient dimension, interpreted as high-dimensional problems enjoying a ``blessing of dimensionality.''
For example, the simplicial interpolation scheme of \citet{belkin2018overfitting} has excess risk that decays as
$O(2^{-d})$ for ambient dimension $d$, and \citet{rakhlin2018consistency} show that kernel ridgeless regression with the Laplace kernel is inconsistent in any fixed dimension, but with a lower bound on risk that decays with dimension.
We find an explicit result to this effect when studying KR: for Laplace kernels and ReLU NTKs, asymptotic excess mean squared error decays like $\Theta(1/d)$.
These phenomena support our claim that many interpolating methods are \emph{tempered} on real distributions, which have fixed dimension.

\ifarxiv
Though most results of \citet{bartlett2019benign} treat linear regression with dimension that grows with $n$, their Theorem 6 shows that benign overfitting can occur in a static infinite-dimensional setting \textit{if} the eigenvalues of the covariance matrix decay at a particular slow rate, with faster decay than this giving non-benign behavior.
Our Theorem \ref{thm:trichotomy}, which is our main theoretical result, essentially subsumes this: we recover this fragile condition for benign overfitting and also derive spectral conditions under which faster decays give either tempered or catastrophic overfitting.

Our KR setting is similar but slightly different from that of \citet{bartlett2019benign} and merits a brief discussion.
\citet{bartlett2019benign} study infinite-dimensional linear regression with sub-Gaussian features and a fixed covariance spectrum, and proceed to obtain upper and lower bounds on test risk that permit one to determine whether fitting is benign.
This setting is mathematically equivalent to KR with random sub-Gaussian eigenfunctions.
Our setting is essentially KR with random Gaussian eigenfunctions.
Instead of studying KR generalization from scratch, we simply use recent results using methods from statistical mechanics to derive fairly explicit closed-form expressions for the expected test MSE of KR with Gaussian features in terms of the kernel spectrum \citep{bordelon:2020-learning-curves, canatar:2021-spectral-bias, jacot:2020-KARE, simon:2021-eigenlearning}.
The result which we use as our starting point is rather simple, and our proofs are consequently quite straightforward and extensible.

Our Theorem \ref{thm:trichotomy} includes the surprisingly simple result that, for KR with a kernel with a powerlaw eigenspectrum with exponent $\alpha$, the kernel overfits target noise by a factor of $(\alpha - 1)$ as $n \rightarrow \infty$, implying that the method is tempered.
The fact that this overfitting factor is $\Theta(1)$ can be obtained by assembling several results of \citet{jacot:2020-KARE}, though their analysis does not provide the value of the constant.
The asymptotic behavior of powerlaw KR was also studied by \citet{spigler:2020}, though they assume zero noise, while noise is central to the present work.
\citet{cui:2021-krr-decay-rates} also study the asymptotics of powerlaw KR and find as we do that, when the ridge parameter is negligible, test risk plateaus as $n \rightarrow \infty$, though they also do not find the value of the asymptotic risk.
\fi

\section{The Three Types of Overfitting}
\label{sec:prelims}

Here we formally present our taxonomy of overfitting, delineating 
the three types of asymptotic behaviors which learning procedures can exhibit.

\subsection{Definitions}
\label{sec:prelim_notation}
We consider a fairly generic in-distribution supervised learning setting. For simplicity, we present
definitions for regression, but these are readily extended to classification.
We wish to learn a function $\hat{f}: \cX \to \R$ from a size-$n$ dataset of i.i.d. samples $\mathcal{D}_n \equiv \{(x_i, y_i)\}_{i=1}^n \sim \mathcal{D}$, where $\D$ is a joint distribution over $\cX \times \R$,
and $\cX$ is the input domain.
We shall generally assume nonzero target noise, with
$\text{Var} [y_i | x_i] > 0$.
We evaluate the generalization performance of $\hat{f}$ by the \emph{mean squared error} (MSE):
$R(\hat{f}) := \E_{x,y\sim\D} \left[(\hat{f}(x) - y)^2\right]$.
The Bayes-optimal regression function is given by $f^* := \argmin_{f} R(f)$, where the minimization is over all measurable functions, and has risk $R^*$ which is called the \emph{irreducible risk}.
The \emph{excess risk} of any function $\hat{f}$ is given by $\bar{R}(\hat{f}) := R(\hat{f}) - R^*$.
We say an estimator achieves \textit{interpolation} if $\wh{f}_n(x_i)=y_i$ for all $(x_i,y_i) \in \D_n$.
These definitions are readily generalized to classification, using classification error in place of MSE.

\paragraph{Learning Procedure.}
The objects of study in our taxonomy are \emph{learning procedures}.
Our definition of a learning procedure is quite general, allowing discussion of methods from DNNs to 1-nearest-neighbors.
Informally, a learning procedure is simply a specification of which model to output on a given train set of a given size.

Formally, a \textit{learning procedure} $\mc A:=\{A_n\}_n$ is a sequence of (potentially stochastic) functions, indexed by sample size $n \in \mathbb{N}$.
At each $n$, the function $A_n: \mathcal{D}_n \mapsto \hat{f}_n$ inputs a train set
$\cD_n$ and outputs a ``model'' $\hat{f}_n: \cX \to \R$.
Note that this $n$-dependence allows learning procedures to be \emph{non-uniform},
varying the learning algorithm with sample size $n$. For example,
it allows procedures which scale up a DNN or narrow a kernel bandwidth as $n$ grows.
The \textit{expected risk} of a learning procedure on $n$ examples from distribution $\cD$
is $\cR_n := \E_{A_n, \D_n} [R(A_n(\D_n))]$.

\subsection{The Taxonomy}

We shall categorize learning procedures in terms of their \textit{asymptotic expected risk}.
We handle regression and $K$-class classification settings separately, due to their different loss scalings.
As the number of samples $n \to \infty$, the sequence of expected
risks $\{\cR_n\}_n$ can behave in three different ways, as listed in Table~\ref{tab:taxonomy}.
These three limiting behaviors define our taxonomy.

\begin{table}[h!]
\centering
\begin{tabular}{@{}lll@{}}
                      & \textbf{Regression} & \textbf{Classification} \\ \toprule
\textbf{Benign}       &  $\lim_{n\rightarrow\infty} \cR_n = R^*$ 
& $\lim_{n\rightarrow\infty} \cR_n = R^*$             
\\ \midrule
\textbf{Tempered}     & $\lim_{n\rightarrow\infty} \cR_n \in (R^*, \infty)$ 
&   
$\lim_{n\rightarrow\infty} \cR_n \in (R^*, 1-\frac{1}{K})$ 
\\ \midrule
\textbf{Catastrophic} &  $\lim_{n\rightarrow\infty} \cR_n = \infty$  
& $\lim_{n\rightarrow\infty} \cR_n = 1-\frac{1}{K}$ 
\\ \bottomrule
\end{tabular}%
\vspace{2mm}
\caption{{\bf Our taxonomy of (over)fitting.}}
\label{tab:taxonomy}
\end{table}

There is technically a fourth option -- that the limit does not exist --
but, to our knowledge, this does not describe any non-pathological algorithms.
All together, this set of behaviors is exhaustive, describing every possible learning procedure.
Note that the definitions for classification and regression are identical,
except for the bounds at $\infty$ replaced by the error of the predictor choosing a
uniformly\footnote{We assume balanced classes throughout, for notational simplicity. The definitions can be modified appropriately for imbalanced classes.} random label ($1-\frac{1}{K}$).

\begin{figure}[t!]
\centering
\begin{subfigure}{1.0\textwidth}
  \centering
    \hspace{1em}
  \includegraphics[width=0.9\linewidth]{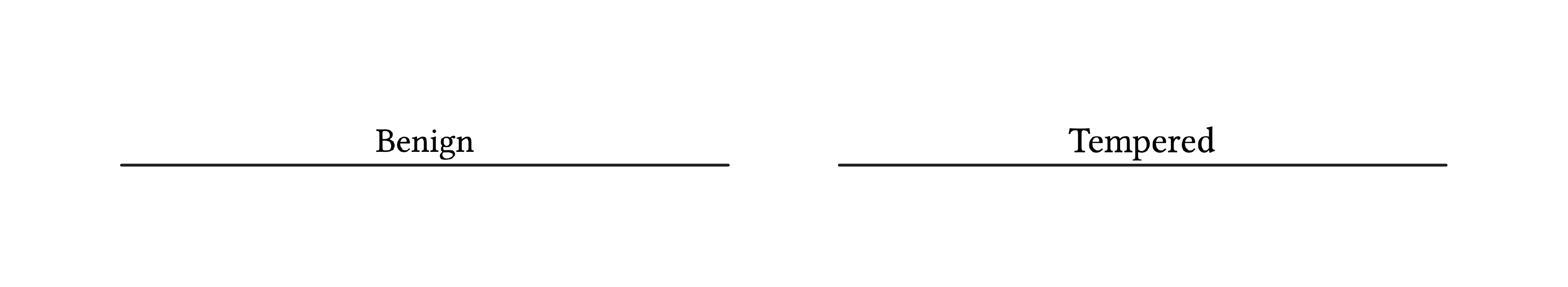}
\end{subfigure}
\begin{subfigure}{.25\textwidth}
  \centering
  \includegraphics[width=1.0\linewidth]{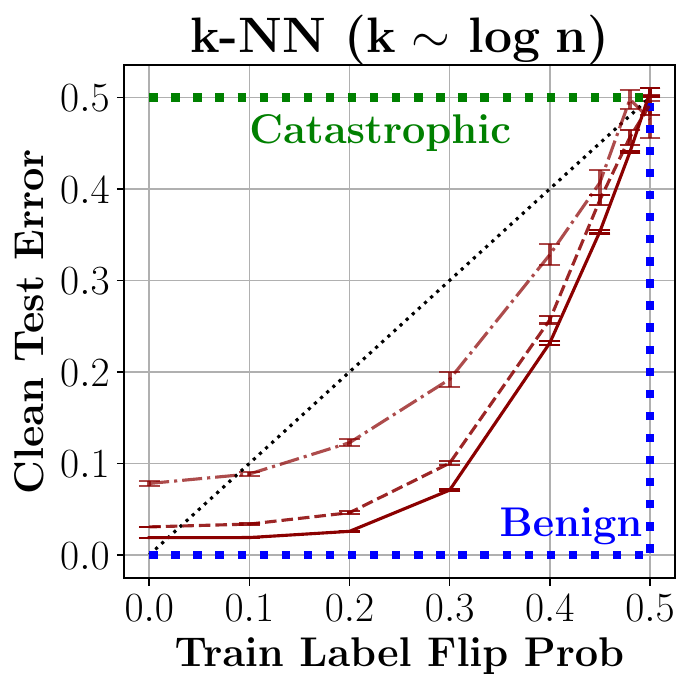}
  \label{fig:binary_mnist_knn_noise_profile}
\end{subfigure}%
\begin{subfigure}{.25\textwidth}
  \centering
  \includegraphics[width=1.0\linewidth]{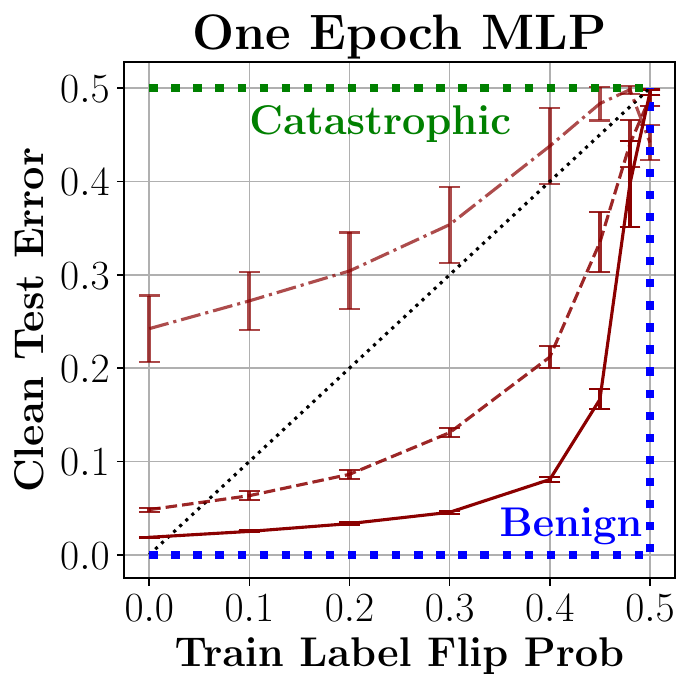}
  \label{fig:binary_mnist_1nn_noise_profile}%
\end{subfigure}%
\begin{subfigure}{.25\textwidth}
  \centering
  \includegraphics[width=1.0\linewidth]{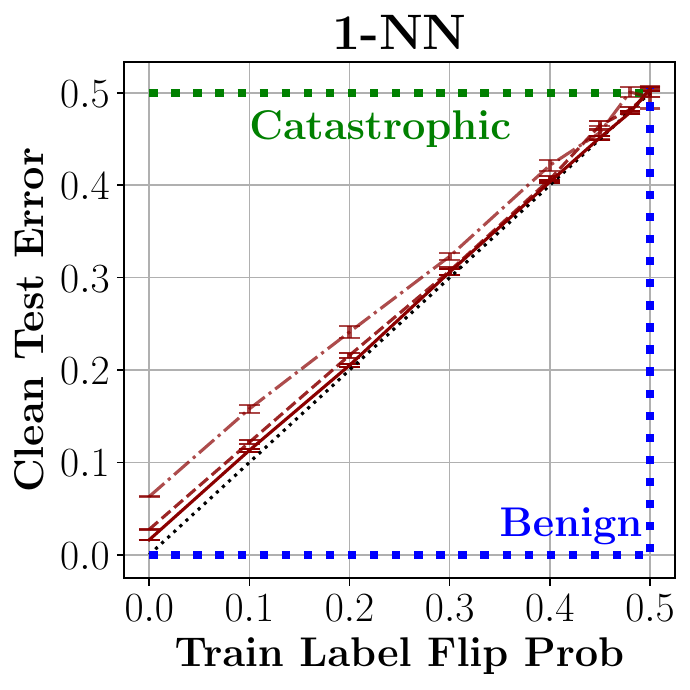}
  \label{fig:binary_mnist_mlp_noise_profile_epoch1}
\end{subfigure}%
\begin{subfigure}{.25\textwidth}
  \centering
  \includegraphics[width=1.0\linewidth]{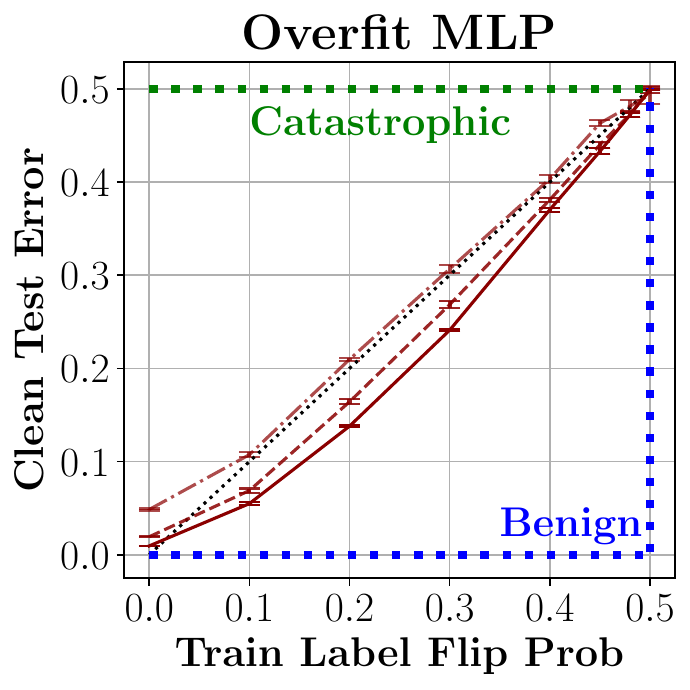}
  \label{fig:binary_mnist_mlp_noise_profile_overfitting}
\end{subfigure}
\begin{subfigure}{1.0\textwidth}
  \centering
  \vspace{-1.0em}
  \includegraphics[width=0.6\linewidth]{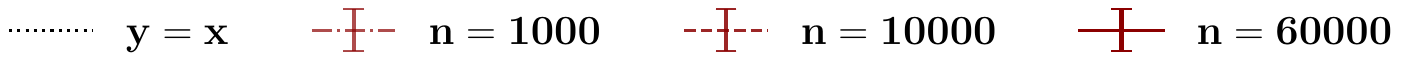}
\end{subfigure}%
\caption{{\bf Examples of Benign and Tempered Fitting.} Noise profiles for several different methods on the Binary-MNIST classification task, showing clean test error as a function of train label noise as the train set $n$ grows.
Left: two methods which
exhibit benign overfitting, with performance converging to Bayes optimal as $n\to \infty$.
Right: two methods which exhibit tempered overfitting, with test error that remains bounded away from $0$.
The two ``tempered'' methods here are interpolating, while the two ``benign'' methods are not. 
Both the benign and the tempered MLP use identical architectures; one is trained for one epoch, and the other trained to interpolation.
Details in Appendix~\ref{apdx:experimental_details}.
}
\label{fig:binary_mnist_mlp_noise_profile}
\end{figure}

\subsection{Noise Profiles}
\label{sec:noiseprofile}

To study asymptotic risk, we use a tool we call a \textit{noise profile}.
A noise profile characterizes the \textit{sensitivity} of a learning procedure to noise in the training set.
For a given learning procedure $\mc A$ and data distribution $\mc D$,
a \emph{noise profile} $\profile_{\mc A}$ describes how the asymptotic risk varies with respect to $\sigma$, the level of artificial noise added to training targets.
Formally, the noise profile $\profile_{\mc A}$ is
\ifarxiv
\begin{equation}
    \profile_{\mc A}(\sigma) = \lim_{n \to \infty} \E_{A_n; \mc D_n(\sigma)} R(A_n(D_n(\sigma))),
\end{equation}
\else
$\profile_{\mc A}(\sigma) = \lim_{n \to \infty} \E_{A_n; \mc D_n(\sigma)} R(A_n(D_n(\sigma)))$,
\fi
where $\cD_n(\sigma)$ denotes $n$ i.i.d. samples from the distribution $\cD$, with $\sigma$-level of label noise added.
The label noise $\sigma \in \R$ denotes a kind of noise that depends on the problem setting: $\sigma$ is the variance of additive Gaussian noise for regression settings,
or the label-flip probability for classification settings.
We cannot empirically evaluate the $n \to \infty$ limit exactly,
so we estimate it using asymptotics from finite but large sample sizes.
As shown in Figure \ref{fig:intro_profile}, noise profiles are easily plotted and reveal at a glance whether an learning procedure is benign, tempered, or catastrophic.

\subsection{Applying Our Taxonomy}

\newcommand{\ours}[1]{\textbf{#1}}
\begin{table}[t]
\setlist[itemize]{leftmargin=*}
\centering
\begin{tabular}{p{0.3\linewidth}p{0.3\linewidth}p{0.3\linewidth}}
Benign & Tempered & Catastrophic \\
\midrule
\begin{itemize}[topsep=0pt]
    \item Early-stopped DNNs
    \item KR with ridge
    \item $k$-NN ($k \sim \log n$)
    \item Nadaraya-Watson kernel \newline smoothing with Hilbert kernel
\end{itemize}
&
\begin{itemize}[topsep=0pt]
    \item \ours{Interpolating DNNs}
    \item \ours{Laplace KR}
    \item \ours{ReLU NTKs}
    \item $k$-NN (constant $k$)
    \item Simplicial interpolation
\end{itemize}
&
\begin{itemize}[topsep=0pt]
    \item \ours{Gaussian KR}
    \item Critically-parameterized \newline regression
\end{itemize}
\end{tabular}
\caption{A taxonomy of models under the three types of fitting identified in this work.
\ours{BOLD} are results from our work, others are known or folklore results.
}
\vspace{-5mm}
\label{tab:fitting_taxonomy}
\end{table}

Our taxonomy handles general learning procedures, even those which do not interpolate their train sets\footnote{We use ``overfitting'' to describe interpolating methods, and ``fitting'' to describe general methods.}.
Thus, we can apply it to describe many existing methods in machine learning.
In doing so, we \emph{refine the language} around statistical consistency:
many methods were known to be statistically inconsistent (i.e. not benign), 
but we highlight that there are two distinct ways to be inconsistent: tempered and catastrophic.

To illustrate our taxonomy, we give several examples of known results in Table~\ref{tab:fitting_taxonomy}.
Any statistically consistent method is by definition \emph{benign}:
this includes non-interpolating methods such as early-stopped DNNs \citep{ji2021early}
and $k$-nearest-neighbors with $k \sim \log n$ \citep{cover1967nearest,chaudhuri2014rates},
as well as interpolating methods such as Nadarya-Watson kernel smoothing \citep{devroye1998hilbert} and k-NN schemes \citep{belkin2018overfitting,belkin2018does}.

Many catastrophic learning procedures are also known.
Parametric models which are ``critically-parameterized'' (i.e. at the double descent peak)
overfit catastrophically: this includes random feature regression
with number of features $p = n$ \citep{mei2019generalization}, and, more generally, for a broad class of linear and random feature methods~\citep{holzmuller2021on}.
Generic empirical-risk-minimization for a hypothesis class with VC-dimension $> n$
will also overfit catastrophically in the worst case.
It turns out that RBF kernel regression under natural assumptions also catastrophically overfits,
as we show in Section \ref{sec:kernel}.

Finally, many methods, from classical to modern, exhibit tempered overfitting.
First, it is well-known that 1-nearest-neighbors (1-NN) converges to an asymptotic risk that is finite but bounded away from Bayes optimal\footnote{In fact,
the excess test MSE converges to the variance of the observation noise.}, which is tempered behavior.
Further, any ``underfitting'' method, such as empirical risk minimization
over a hypothesis class with VC-dimension $\ll n$, will be tempered if the
hypothesis class does not contain the ground-truth function.
In Section \ref{sec:experiments}, we empirically demonstrate that many standard DNNs exhibit 
tempered overfitting when trained to interpolation.
We complement this with theoretical results in Section \ref{sec:kernel},
showing that kernel regression with the Laplace kernel,
as well as with the ReLU NTK (which describes the training of an infinite-width ReLU fully-connected network \citep{jacot2018neural}), exhibits tempered overfitting.
Our new category of tempered overfitting is thus not merely a theoretical possibility
but in fact captures many natural and widely-used learning methods.

In Figure \ref{fig:binary_mnist_mlp_noise_profile} we demonstrate our taxonomy experimentally for two benign methods ($k$-NN and early-stopped MLPs) and two tempered methods (1-NN and interpolating MLPs) 
on a binary classification version of MNIST, with varying noise in the train labels.
We plot test classification error on the clean test set against the proportion of flipped labels
in the training set.
As $n$ grows, the benign methods approach zero test error even at nonzero train noise,
while the tempered methods converge to a test error bounded away from zero.

\section{Overfitting in Kernel Regression}
\label{sec:kernel}

We begin with a study of \textit{kernel regression} (KR), a widely-used nonparameteric learning algorithm which, we will see, is sufficiently rich to exhibit all three regimes of overfitting, yet sufficiently simple that this can be shown analytically.
Theoretical interest in this algorithm has increased significantly in recent years due to the discovery that trained DNNs converge to ridgeless KR in the infinite-width and infinite-time limit \citep{jacot2018neural}, implying that insights into KR simultaneously shed light on overparameterized DNNs.
The overparameterized linear regression setting of \citet{bartlett2019benign} is equivalent to KR, and we will make a direct comparison with their results at the end of the section.

KR is fully specified by a positive-semidefinite kernel function $K : \mathbb{R}^d \times \mathbb{R}^d \rightarrow \mathbb{R}$ and a ridge parameter $\delta \ge 0$.
We allow the training set $\D_n$ to contain $n$ samples $(x_i, y_i) 
\sim \D$, and we assume that $y_i = f^*(x_i) + \eta_i$ with true function $f^*$ and noise $\eta_i \sim \mathcal{N}(0,\sigma^2)$.
KR returns the predicted function $\hat{f}$ given by
\begin{equation} \label{eqn:krr_def}
	\hat{f}(x) = K(x, \D_n) \left( K(\D_n,\D_n) + \delta \mathbf{I}_n \right)^{-1} \mathcal{Y},
\end{equation}
where $K(\D_n,\D_n)$ is the ``data-data kernel matrix'' with components $K(\D_n,\D_n)_{ij} = K(x_i, x_j)$, $K(x,\D)$ is a row vector with components $K(x,\D)_i = K(x, x_i)$, and $\mathcal{Y}$ is a column vector of
\ifarxiv
target labels\footnote{We note that some works use $n \delta$ instead of $\delta$ in Equation \ref{eqn:krr_def}, a convention which gives an asymptotically-biased estimator and changes some downstream conclusions about KR with ridge.}.
\else
targets.
\fi

Existing literature provides examples of KR exhibiting all three asymptotic behaviors.
Benign overfitting has been analyzed for overparameterized linear regression \citep{advani2017high,bartlett2019benign, muthukumar2020harmless, hastie2019surprises, belkin2019two}, a special case of KR.
It is well-known that KR with a positive ridge value (with appropriate scaling conventions) is consistent and thus benign \citep{christmann:2007-krr-consistency}.
Furthermore, in 1D, a Laplace kernel approaches piecewise linear interpolation as $n \rightarrow \infty$ \citep{belkin2018overfitting}, which is easily shown to exhibit tempered overfitting, and \citet{rakhlin2018consistency} proved that the Laplace kernel does not overfit benignly in (fixed) dimension greater than one (though did not say whether it was in fact tempered or catastrophic).
Finally, it is known in experimental folklore (though not theoretically, to our knowledge) that KR with a Gaussian kernel and zero ridge tends to yield poorly-conditioned kernel matrices and catastrophic behavior.
Here we derive fairly general conditions under which KR falls into each regime, solidifying these various observations into a unified picture.

As our chief tool for obtaining these conditions, we use a recently-derived closed-form approximation for the expected test MSE of KR \citep{bordelon:2020-learning-curves, canatar:2021-spectral-bias, jacot:2020-KARE, simon:2021-eigenlearning}.
By simply taking the $n \rightarrow \infty$ limit of this expression, we can classify a given kernel into one of our three regimes.

The expression we will use gives test MSE in terms of the eigenspectrum of the kernel (as given by the Mercer decomposition) and the eigendecomposition of the target function.
The nonnegative eigenvalues $\lambda_1 \ge \lambda_2 \ge ... \ge 0$ and orthonormal eigenfunctions $\{ \phi_i \}_{i=1}^\infty$ are given by
\begin{equation}
	\mathbb{E}_{x' \sim p} [ K(x, x') \phi_i(x') ] = \lambda_i \phi_i(x),  \ \ \ \ \  \text{where} \ \ \ \  \mathbb{E}_{x \sim p} [ \phi_i(x) \phi_j(x) ] = \delta_{ij}.
\end{equation}
We note that a kernel must be positive semidefinite and that $\sum_i \lambda_i = \text{Tr}[K] = \mathbb{E}_{x \sim p} [ K(x,x) ]$, which we assume is finite.
Because the eigenfunctions form a complete basis, we are free to decompose the target function as $f^*(x) = \sum_i v_i \phi_i(x)$, where $\{ v_i \}_{i=1}^\infty$ are eigencoefficients.

The above-mentioned works derive equivalent closed-form expressions for the test MSE of KR in terms of this spectral information using methods from the statistical physics
\ifarxiv
literature\footnote{
Specifically, \citet{bordelon:2020-learning-curves} use a PDE approximation, \citet{canatar:2021-spectral-bias} use a replica calculation, \citet{jacot:2020-KARE} use tools from random matrix theory, and \citet{simon:2021-eigenlearning} use a calculation inspired by the cavity method.
}.
\else
literature.
\fi
These methods are nonrigorous and rely on approximations (see Appendix \ref{app:kr_proofs} for a discussion), but they are expected to become exact in the large-$n$ limit, and comparison with empirical KR generally confirms a close match even at modest $n$.
Here we use the framework of \citet{simon:2021-eigenlearning}, which expresses the final result in terms of ``modewise learnabilities" $\{\cL_i\}_{i=1}^\infty$, a set of scores in $[0,1]$ which indicate how well each eigenmode is learned at a given $n$.
This choice of variables will simplify our proofs.

\citet{simon:2021-eigenlearning} find that test MSE $\cR_n$ is approximated by
\begin{equation} \label{eqn:eigenlearning}
    \begin{split}
        \cR_n &\approx \e_n \equiv \e_0 \left( \sum_i (1 - \cL_i)^2 v_i^2 + \sigma^2 \right),
        \ \ \ \ \text{where} \ \ \ \
        \e_0 \equiv \frac{n}{n - \sum_j \cL_j^2}, \\
        \cL_i &\equiv \frac{\lambda_i}{\lambda_i + \C},
        \ \ \ \ \text{and} \ \ \ \
        \C \ge 0 \text{ satisfies } \sum_i \frac{\lambda_i}{\lambda_i + \C} + \frac{\delta}{\C} = n.
    \end{split}
\end{equation}
This MSE includes noise on test labels; to instead compute a value for \textit{excess} risk, one would simply subtract $\sigma^2$.

Here we study the asymptotic behavior of $\e_n$ as $n \rightarrow \infty$ for varying eigenspectra and ridge values.
The fitting regime of the kernel is then given by this limit:
(a) if $\lim_{n \rightarrow \infty} \e_n = \sigma^2$ (the Bayes-optimal MSE), then fitting is \textit{benign},
(b) if $\lim_{n \rightarrow \infty} \e_n \in (\sigma^2, \infty)$, then fitting is \textit{tempered}, and
(c) if $\lim_{n \rightarrow \infty} \e_n = \infty$, then fitting is \textit{catastrophic}.
We obtain conditions on the kernel eigenspectrum under which KR falls into each of these three regimes.

Our proofs rely on several (quite weak) technical assumptions on $\{\lambda_i\}_i$ and $\{v_i\}_i$, the most important of which is that the target function does not place weight in zero-eigenvalue modes (i.e. outside the kernel's RKHS).
We defer enumeration and discussion of these conditions to Appendix \ref{app:kr_proofs}, where they are listed as Assumption \ref{assumption:vals_and_coeffs}. Our result is the following:
\begin{theorem}[KR trichotomy] \label{thm:trichotomy}
For $\{ \lambda_i \}_{i=1}^\infty$ and $\{v_i\}_{i=1}^\infty$ satisfying Assumption \ref{assumption:vals_and_coeffs}, $\sigma^2 > 0$, and $\e_n$ given by Eq. \ref{eqn:eigenlearning_repr},
\begin{enumerate}[(a)]
\item\label{thm:case:delta_positive} If $\delta > 0$ or $\lambda_i = i^{-1} \log^{-\alpha} i$ for some $\alpha > 1$,
then $\lim\limits_{n\rightarrow\infty} \e_n = \sigma^2$.
\vspace{-2mm}
\item\label{thm:case:delta_zero_polytail} If $\delta = 0$ and $\lambda_i = i^{-\alpha}$ for some $\alpha > 1$, then $\lim\limits_{n\rightarrow\infty} \e_n = \alpha \sigma^2$.
\vspace{-2mm}
\item\label{thm:case:delta_zero_exptail} If $\delta = 0$ and $\lambda_i = i^{-\log i}$, or more generally if $\frac{\lambda_i}{\lambda_{i+1}} \ge \frac{i^{-\log i}}{(i+1)^{-\log (i+1)}}$ for all $i$, then $\lim\limits_{n\rightarrow\infty} \e_n = \infty$.
\end{enumerate}
\end{theorem}

We defer the proof to Appendix \ref{app:kr_proofs}.
The proof proceeds by first showing that asymptotic MSE is dominated by the noise, not the true function, and then computing $\lim_{n\rightarrow\infty}\e_0$ in each of the three cases.

Theorem \ref{thm:trichotomy} can be summarized as follows: a ridge parameter or extremely slow eigendecay leads to benign fitting, powerlaw decay of eigenvalues leads to tempered overfitting, and eigenvalue decay at least as fast as $i^{- \log i}$ leads to catastrophic overfitting.
This theorem strongly suggests the satisfying heuristic that decay \textit{slower} than any $\alpha > 1$ powerlaw is benign, powerlaw decay itself is tempered, and decay \textit{faster} than any powerlaw is catastrophic\footnote{
We leave a complete proof of this heuristic as an open problem.
}.
As $\alpha$ grows in $(1,\infty)$, powerlaw spectra fully interpolate between benign and catastrophic, suggesting we have not missed any regime of interest to our taxonomy.
The fact that the asymptotic MSE from a powerlaw spectrum is simply $\alpha \sigma^2$ is a pleasant
\ifarxiv
surprise\footnote{In our proof, this simple result comes from a cancellation of fairly complex terms, suggesting there exists a simpler way to obtain it.}.
\else
surprise.
\fi

Theorem \ref{thm:trichotomy} has several consequences for KR with familiar kernels and the training of infinite-width networks.
For illustrative purposes, we contrast the Gaussian (RBF) kernel $K_G(x_1, x_2) = e^{-w^{-2} |\!|x_1 - x_2|\!|_2^2}$ with the Laplace kernel $K_L(x_1, x_2) = e^{-w^{-1} |\!|x_1 - x_2|\!|_2}$, where $w$ is a bandwidth parameter.
With data drawn from a $d$-dimensional manifold, the Gaussian and Laplace kernels have eigenspectra that decay like $\lambda_i \sim e^{-i^{2/d}}$ (as can be seen by taking a Fourier transform of $K_G$) and $\lambda_i \sim i^{-(d+1)/d}$ \citep{bietti:2020-deep-equals-shallow},
\ifarxiv
respectively\footnote{At large $n$, only these decaying eigenvalue tails matter for generalization.}.
\else
respectively.
\fi
We note that ReLU NTKs, restricted to the hypersphere, have the same eigendecay as the Laplace kernel \citep{geifman:2020-laplace-ntk, chen:2020-laplace-ntk}.
The implications of Theorem \ref{thm:trichotomy} include the following:

\begin{itemize}
    \item KR with a fixed positive ridge parameter will fit any function in the kernel's RKHS benignly as $n \rightarrow \infty$.
    \item Ridgeless KR with \textit{Laplace kernels} or \textit{ReLU NTKs} will exhibit \textit{tempered overfitting} with $\Theta(1/d)$ excess MSE, approaching benignness as dimension
    \ifarxiv
    grows\footnote{This improves upon the exponential lower bound for the asymptotic MSE of Laplace KR in \citet{rakhlin2018consistency}.}.
    \else
    grows.
    \fi
    \item Ridgeless KR with the \textit{Gaussian kernel} will exhibit \textit{catastrophic overfitting}.
    \item The asymptotic behavior of KR with non-ReLU NTKs \textit{depends on the activation function}.
    Virtually any kernel on the $d$-sphere (including the Gaussian kernel) can be realized as the NTK of a wide network with a proper choice of activation function \citep{simon:2022-rev-eng}, and thus there exist choices of activation function that yield catastrophic as well as tempered overfitting.
    \item Early stopping is known to act as an effective ridge parameter for wide networks \citep{ali:2019-effective-ridge}, and thus we should expect that early-stopped wide networks will fit benignly.
\end{itemize}

We provide experiments illustrating Theorem \ref{thm:trichotomy} with Gaussian and Laplace kernels in Section \ref{sec:experiments}.
We additionally check Theorem \ref{thm:trichotomy}b with several subsequent experiments:
Figure \ref{fig:powerlaw_kr} shows that, in synthetic KR with Gaussian eigenfunctions and exact powerlaw spectra, $\lim_{n \rightarrow \infty} \e_n$ is $\alpha \sigma^2$.
And Figure \ref{fig:laplace_kr} in Appendix \ref{app:powerlaws}
shows that, as predicted, Laplace kernels indeed appear to have asymptotic risk that decays like $\Theta(1/d)$.

We note that, to our knowledge, no well-known kernel has an eigendecay slower than all $\alpha > 1$ powerlaws in finite dimension, and finding one in closed form --- yielding ridgeless KR which overfits benignly --- is an interesting open problem.

\ifarxiv
\paragraph{Comparison with \citet{bartlett2019benign}.}
Our setting is essentially the same as the overparameterized linear regression setup studied by \citet{bartlett2019benign}:
KR is equivalent to linear regression in eigenfeature space, and our kernel eigenvalues are equivalent to their covariance matrix eigenvalues.
Our spectral condition for benignness in Theorem \ref{thm:trichotomy}a is precisely that of their Theorem 6.1, and we include it for completeness.
Their Theorem 6.1 notes that powerlaw decays with $\alpha > 1$ are \textit{not benign}; here we find that they are in fact \textit{tempered}.
\fi

\begin{figure}[t!]
\centering
\begin{subfigure}{1\textwidth}
  \centering
  \includegraphics[width=0.95\linewidth]{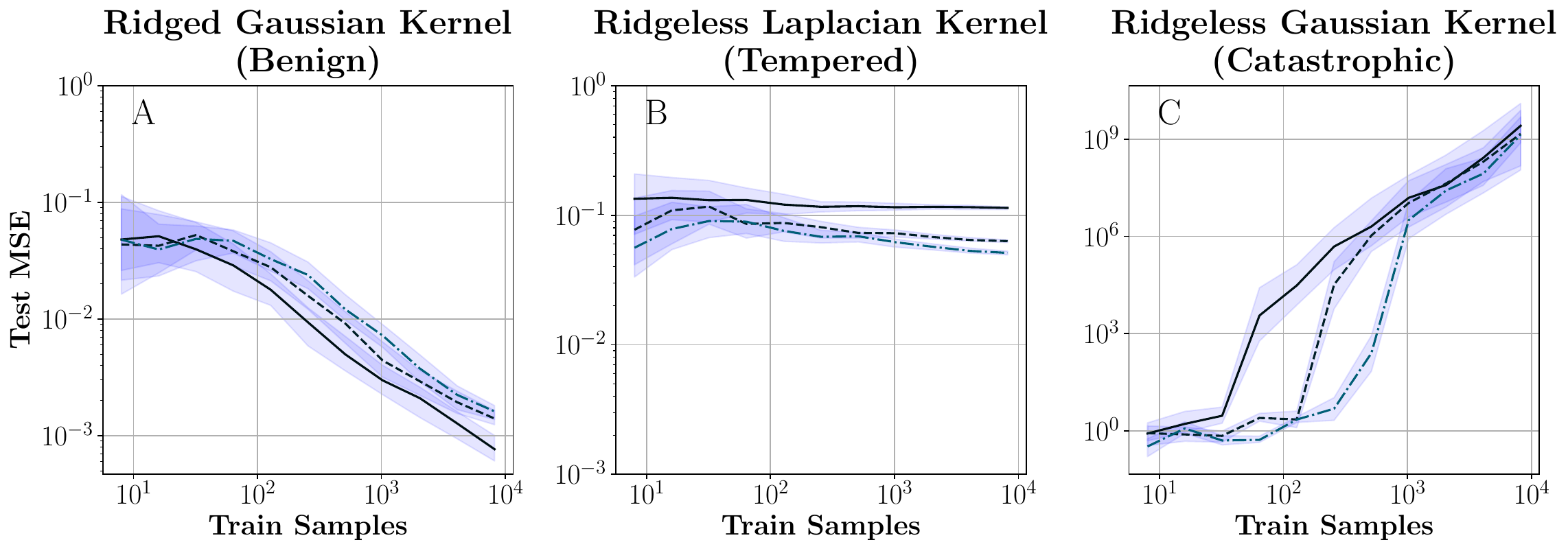}
\end{subfigure}
\begin{subfigure}{0.5\textwidth}
  \centering
  \includegraphics[width=0.9\linewidth]{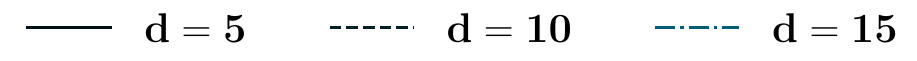}
\end{subfigure}
\caption{
\textbf{Kernel regression can exhibit all three fitting regimes with proper choice of ridge parameter and kernel.}
Plots show learning curves for KR with data $\{x_i\}$ sampled uniformly from the unit sphere $\cS^{d-1}$, trained with pure noise target labels $y_i \sim \mathcal{N}(0,1)$.
Test MSE is computed with respect to a clean test set.
\textbf{(a)} KR with a Gaussian kernel and nonzero ridge is asymptotically benign.
A ridge value of $\delta = 0.1$ was used.
\textbf{(b)} Ridgeless KR with a Laplace kernel exhibits tempered overfitting.
\textbf{(c)} Ridgeless KR with a Gaussian kernel exhibits catastrophic overfitting.
}
\label{fig:kernel_trichotomy_plots}
\vspace{-3mm}
\end{figure}

\section{Experiments}
\label{sec:experiments}

Having demonstrated the three types of fitting theoretically in KR,
we now present a series of experimental results illustrating these regimes in both KR and deep neural networks (DNNs).
\ifarxiv
We show results on various common datasets and models to give evidence that these
phenomena are not specific to our particular theoretical setting,
but are widely relevant across machine learning.
First, we empirically verify Theorem \ref{thm:trichotomy} through simple synthetic experiments with ridged kernels, and the ridgeless Laplace and Gaussian kernel.
We see these three choices of kernels exhibit the three different regimes of overfitting,
exactly as predicted by our theorem.
This confirms both that our theoretical heuristics (which were nonrigorous in places) are empirically accurate, and that the asymptotic behavior is evident at realistically large sample sizes.

We then perform similar regression and classification experiments using MLPs, to show that, beyond the kernel setting, simple DNNs can also exhibit both benign and tempered overfitting ---
depending on the early-stopping criteria.
\fi
We provide full experimental details in Appendix \ref{apdx:experiments}.

\subsection{Experiments on Kernel Regression}
In Figure \ref{fig:kernel_trichotomy_plots}, we run KR
on the following synthetic data distribution:
the inputs $x$ are sampled from the unit sphere $\cS^{d-1}$,
and the targets $y$ are zero-mean Gaussian noise ($y \sim \mathcal{N}(0, 1)$).
This is an extremely simple regression setting: we are just trying to learn the constant-$0$
function under Gaussian observation noise.
We run KR with three choices of kernel: (A) Gaussian kernel with ridge, (B) Laplace kernel without ridge, and (C) Gaussian kernel without ridge.
Figure \ref{fig:kernel_trichotomy_plots} shows that as we increase the sample size,
these three settings exhibit benign, tempered, and catastrophic behavior.
This agrees with the spectral predictions of Theorem \ref{thm:trichotomy}\footnote{Though not reported here, we find the ridged Laplace kernel also exhibits benign fitting as expected.}.

\ifarxiv
Interestingly, Figure \ref{fig:kernel_trichotomy_plots}c shows that increasing input dimension tends to both improve MSE at fixed sample size $n$, and increase the ``critical'' $n$ at which MSE begins to diverge.
This suggests that high dimensionality can be a ``blessing'' which 
prevents catastrophic methods from diverging, agreeing
with the spirit of prior work on the blessings of dimensionality for interpolating methods (e.g. \citep{rakhlin2018consistency}).
\fi

\ifarxiv
\begin{figure}[t]
    \centering
    \includegraphics[width=1.0\textwidth]{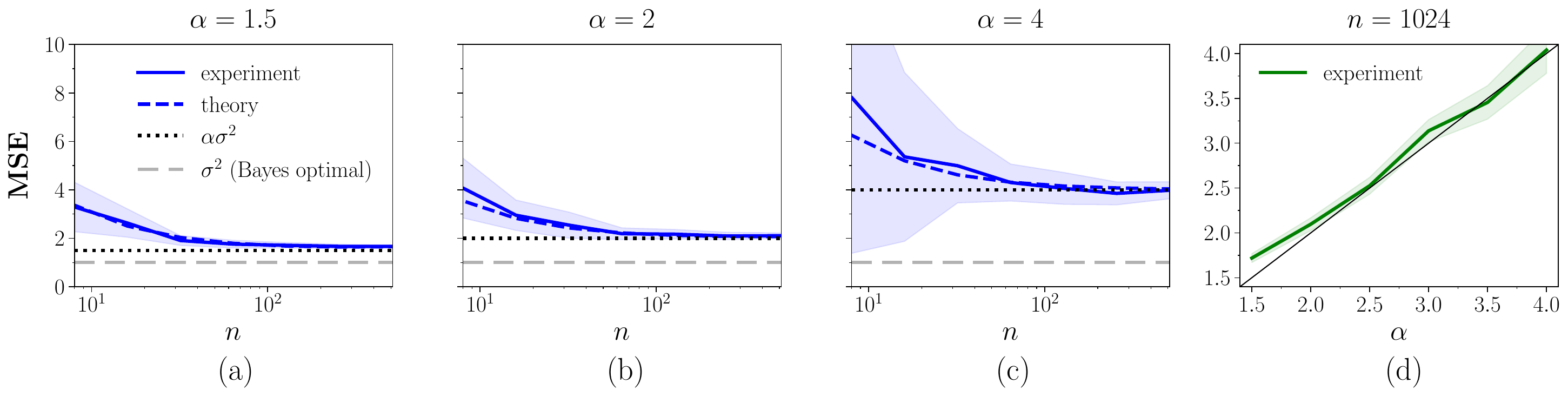}
    \caption{
    \textbf{As $n$ grows, the MSE of KR with Gaussian eigenfunctions and powerlaw kernel eigenspectra with exponent $\alpha$ approaches $\alpha \sigma^2$.}
    \textbf{(a-c)}: learning curves with different $\alpha$.
    \textbf{(d)}: test MSE at $n = 1024$ for varying $\alpha$, with the identity function shown by the solid line.}
    \label{fig:powerlaw_kr}
\end{figure}

In Figure \ref{fig:powerlaw_kr} we explore in more detail the dependency between a powerlaw kernel's spectral decay rate and its asymptotic risk.
We follow the same train and test setup for these experiments, but construct special kernels
such that their spectrum is precisely a powerlaw with exponent $\alpha$ and their eigenfunctions are indeed random Gaussians.
This experimental setup, described in full in Appendix \ref{app:powerlaws}, is equivalent to linear regression with random Gaussian covariates and a covariance kernel with known spectrum.
We see that as the spectral decay $\alpha$ varies,
the (asymptotic) test MSE varies as approximately $\alpha\sigma^2$, agreeing with part (b) of Theorem~\ref{thm:trichotomy}.
\else
In Appendix \ref{app:powerlaws}, we report KR experiments using synthetic kernels with exact powerlaw spectra trained on noisy data.
We find that as the spectral decay $\alpha$ varies,
the (asymptotic) test MSE is indeed approximately $\alpha\sigma^2$ in agreement with part (b) of Theorem~\ref{thm:trichotomy}.    
\fi

\subsection{Experiments on Deep Neural Networks}

\ifarxiv
\begin{figure}[t]
     \centering
     \begin{subfigure}[b]{0.4\textwidth}
     \centering
    \includegraphics[width=0.93\textwidth]{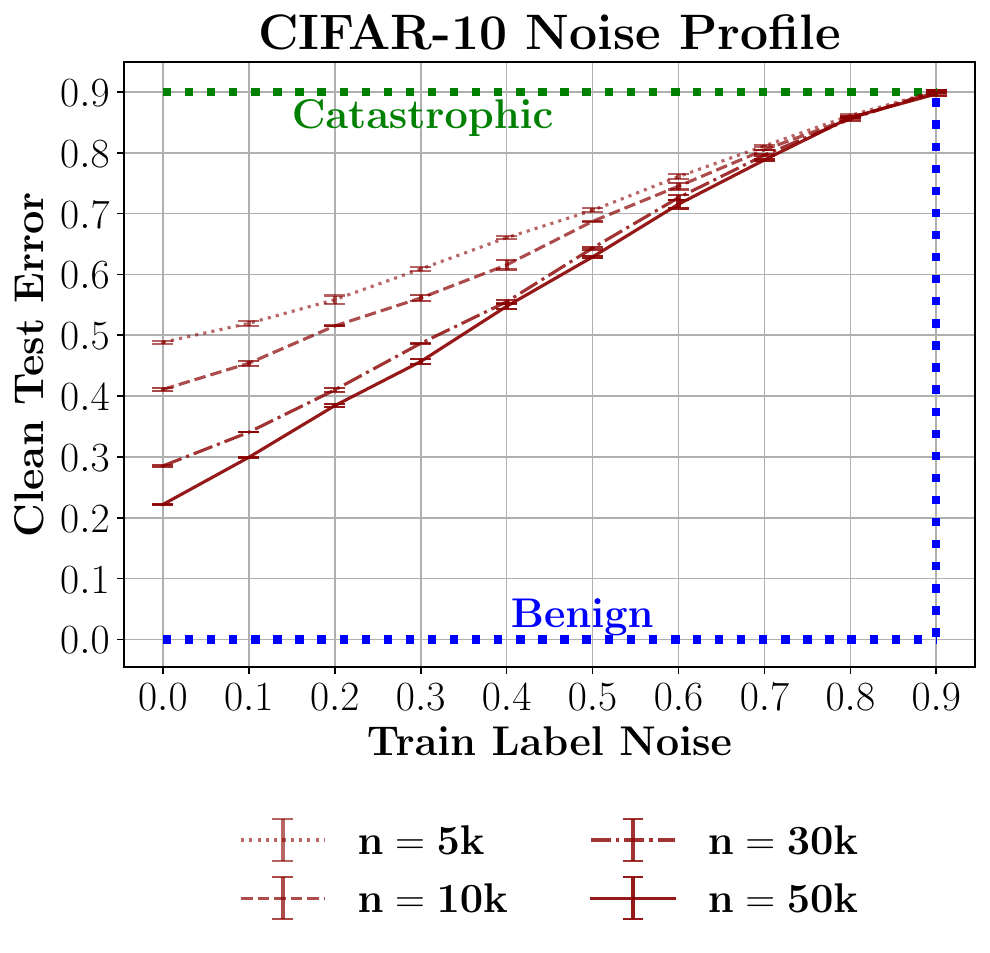}
    \caption{\label{fig:profile_wideresnet_cifar10} WideResNet trained on CIFAR-10.
    }
     \end{subfigure}
    \hspace{2em}
    \begin{subfigure}[b]{0.4\textwidth}
    \centering
    \includegraphics[width=\textwidth]{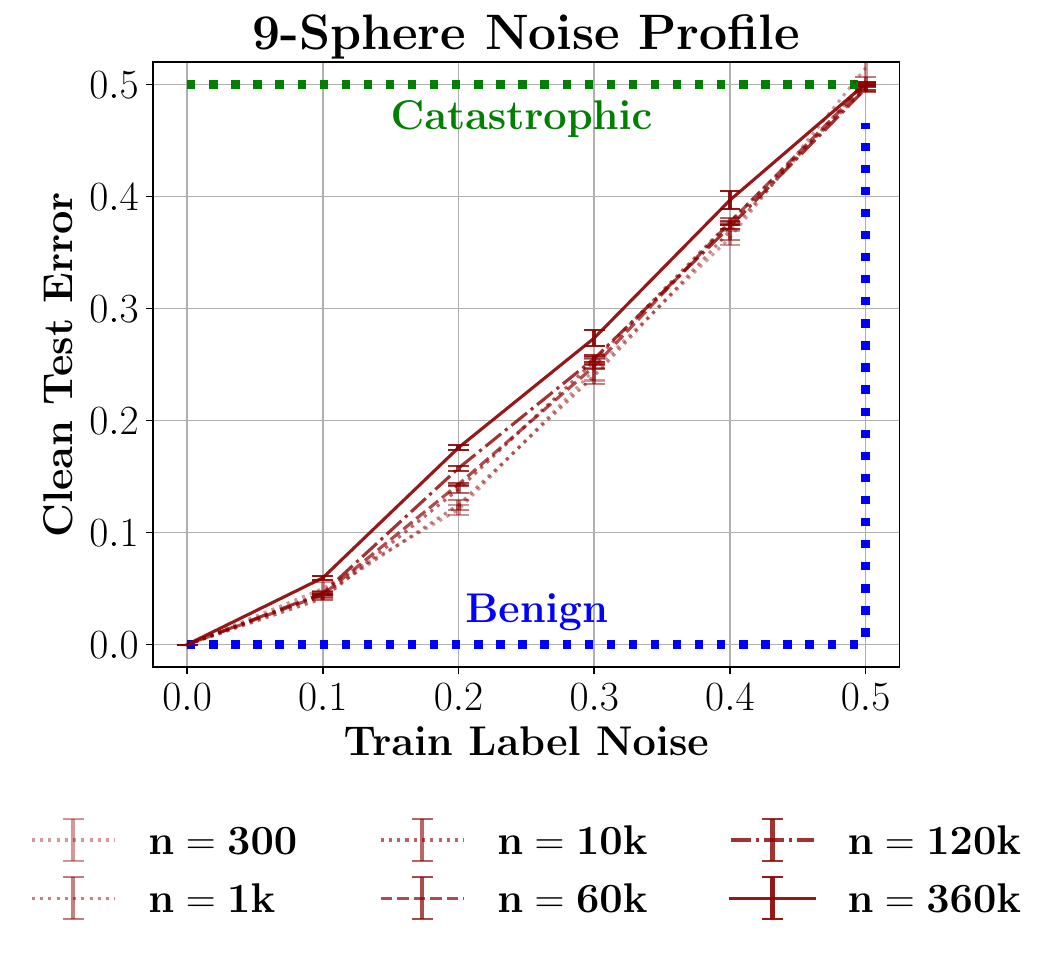}
    \caption{MLP trained on the constant function.}
    \label{fig:synthetic_noise_profile}
     \end{subfigure}
     \caption{\textbf{Noise profiles for interpolating DNNs on CIFAR-10 and synthetic classification tasks.}
     In both settings, the noise profiles asymptotically behave as \emph{tempered} overfitting: neither catastrophic nor benign.}
\end{figure}

\begin{figure}
    \centering
    \begin{subfigure}{.45\textwidth}
      \centering
        \includegraphics[width=0.8\linewidth]{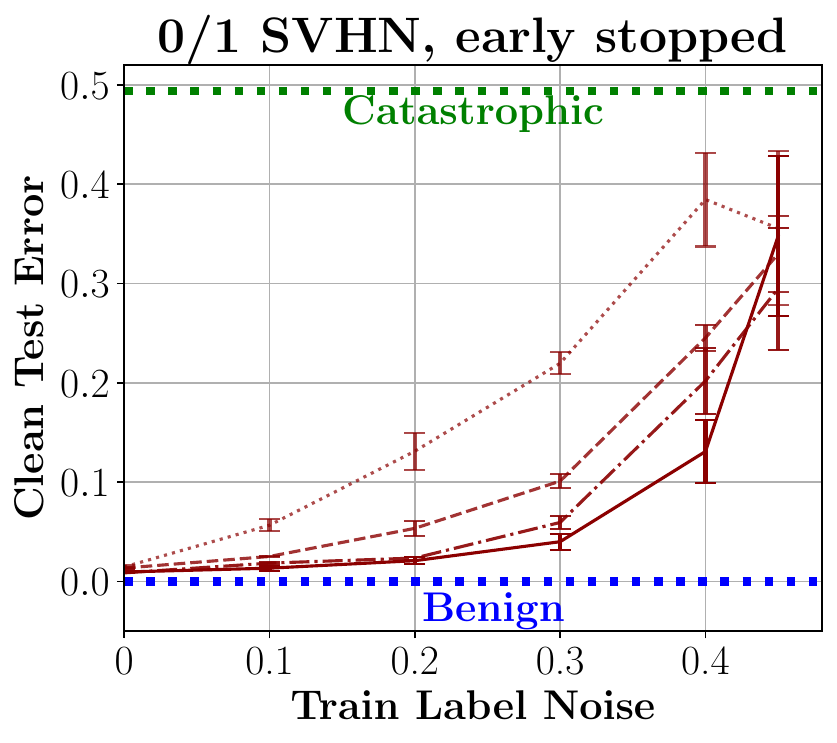}
      \label{fig:svhn_early_stopped}
    \end{subfigure}%
    \begin{subfigure}{.45\textwidth}
      \centering
        \includegraphics[width=0.8\linewidth]{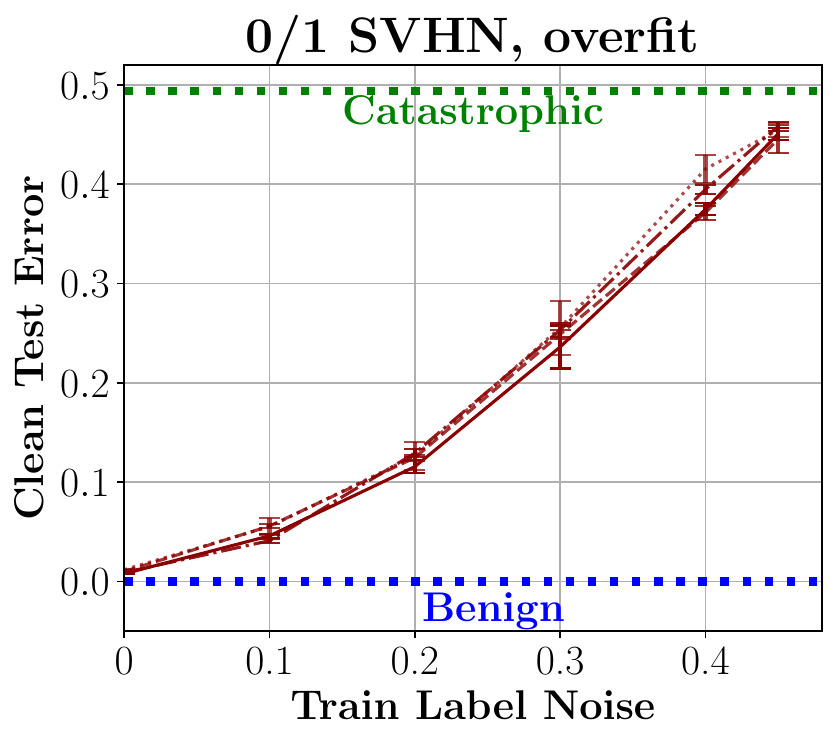}
      \label{fig:svhn_overfit}
    \end{subfigure}
    \begin{subfigure}{1.0\textwidth}
      \centering
      \includegraphics[width=0.6\linewidth]{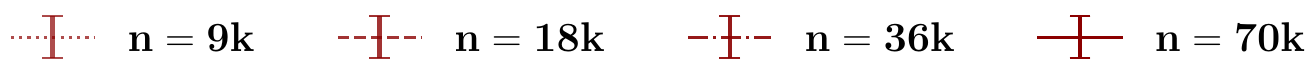}
    \end{subfigure}%
    \caption{
    \textbf{Early-stopped DNNs exhibit fairly benign fitting, while DNNs trained to interpolation exhibit tempered overfitting.}
    We show noise profiles for ResNets trained on Binary SVHN (0/1 digit classification).
    \textbf{(a)} With early stopping, noise profiles approach zero test error as $n$ increases, even with finite label noise, indicating benign fitting.
    \textbf{(b)} When training to interpolation, noise profiles converge to a limiting, roughly linear curve as $n$ increases, indicating tempered fitting.
    }
    \label{fig:svhn_profile}
\end{figure}
\fi

\paragraph{Interpolating DNNs.}
Figure \ref{fig:intro_profile} and Figure \ref{fig:profile_wideresnet_cifar10} show the noise profiles for ResNets trained to interpolation on \textit{two-class} and \textit{ten-class} versions of CIFAR-10, respectively.
In both settings, interpolating DNNs do not approach Bayes optimality, and instead exhibit tempered overfitting.
This tempered behavior is widespread across even much simpler DNN settings.
For example, in Figure \ref{fig:synthetic_noise_profile}, we 
train a three-layer interpolating MLP for binary classification on an extremely
simple synthetic dataset: with inputs drawn from $\mathcal{S}^{9}$ and ground-truth labels as the constant function
$f^*(x) = 1$.
Even in this simple setting, at large sample size, interpolating DNNs are not benign---they do not
successfully learn the constant function in the presence of even slight label noise.
Note that although this experiment is outside the technical scope of Theorem \ref{thm:trichotomy},
it is heuristically consistent:
wide ReLU MLPs tend to have NTKs with powerlaw spectra,
and thus will exhibit tempered overfitting when trained in the NTK regime.

\paragraph{Early-stopped DNNs.}
We now consider DNNs that have been optimally early-stopped.
Figure \ref{fig:svhn_profile} shows noise profiles for Wide ResNets trained on a binary version of SVHN,
both \textit{stopped early} (Figure \ref{fig:svhn_profile}a) and \textit{trained to interpolation} (Figure \ref{fig:svhn_profile}b).
The early-stopped ResNets approach benign fitting as $n$ grows,
with low error even at sizable noise levels, while the ResNets trained to interpolation quickly converge to a tempered noise profile.
This mirrors the behavior of MLPs on binary MNIST, after one epoch of training and after interpolation,
shown earlier in Figure \ref{fig:binary_mnist_mlp_noise_profile}.
Although these benign DNN results are outside the formal scope of known theoretical results, it is heuristically consistent
with results such as \citet{ji2021early}, which show that certain wide and shallow ReLU MLPs are consistent when early-stopped.

\ifarxiv
{
\begin{figure}[t]
    \centering
    \includegraphics[width=0.55\textwidth]{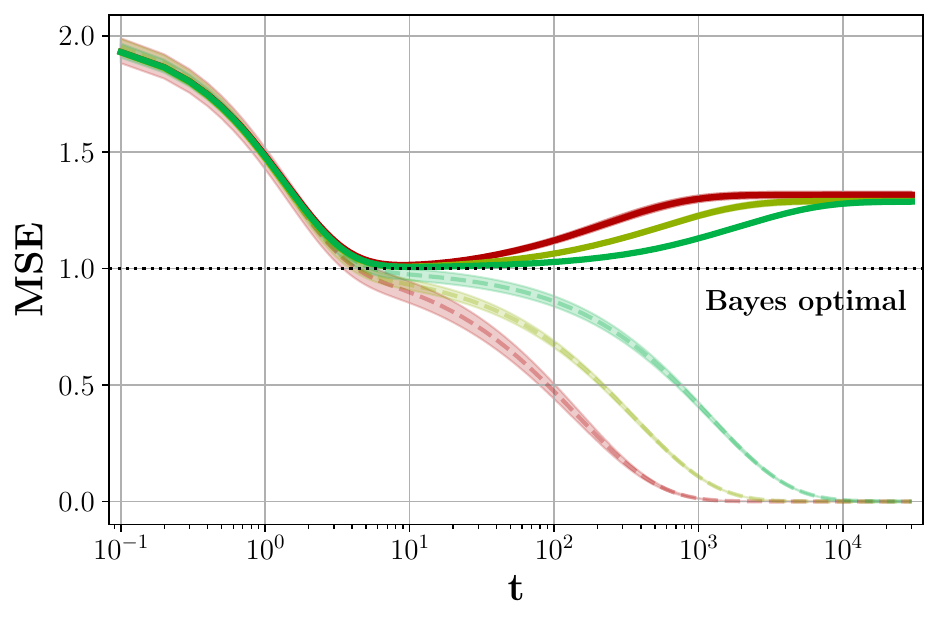}
    \includegraphics[width=0.5\textwidth]{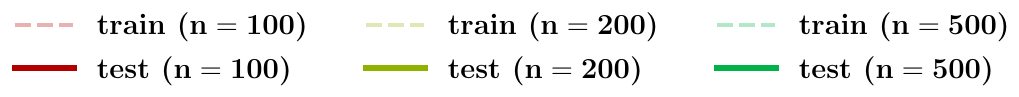}
    \caption[Ba]{\textbf{MLPs trained on noisy data fit \textit{benignly} at early times and exhibit \textit{tempered overfitting} at late times.}
    Curves show train and test MSE for a $4 \times 500$ ReLU MLP with target labels $y_i \sim \mathcal{N}(1,1)$ and varying trainset size $n$.
    Benign fitting, with near-Bayes-optimal test MSE, begins around $t = 3$, ending as the network begins to fit the noise in the training set.
    Tempered overfitting is reflected in the suboptimal-but-finite asymptotic test MSE as $t$ becomes large.
    }
    \label{fig:time_dependence}
\end{figure}

\paragraph{Time dynamics.}
The above discussion suggests that as a single DNN is trained,
it exhibits benign fitting early in training (when it has not fit the noise in its train set), and then transitions to tempered overfitting late in training (as it eventually fits the noise).
In Figure~\ref{fig:time_dependence}, we show a simple experiment that illustrates these time dynamics.
We train a ReLU MLP on the following synthetic regression task (similar to Figure~\ref{fig:synthetic_noise_profile}):
The inputs $x$ are sampled from the unit sphere $\mathcal{S}^4$, and the targets are $\mathcal{N}(1, 1)$.
That is, we are simply trying to learn the constant function $f^*(x) = 1$ on the unit sphere, with Gaussian observation noise.
We then plot the test MSE as a function of the number of optimization steps taken.

From Figure~\ref{fig:time_dependence}, we see that for all train sizes $n$,
the network quickly reaches nearly Bayes-optimal test performance early in training.
In this initial fitting phase, train and test MSE decrease together:
the network is fitting the ground-truth function but has not yet fit the noise.
Once the train MSE drops below the Bayes risk, however, the dynamics change.
At this point (around time $t=4$ in Figure~\ref{fig:time_dependence}),
the network must start to fit noise in the train set, since the train loss drops below the optimal possible test loss.
As training continues, we see the train loss decrease, but the test loss \emph{increase} for the remainder of training: fitting noise in the train set hurts test performance.
Finally, in the limit of large time $t$, the network converges to a test loss that is tempered:
above Bayes-optimal, but asymptotically constant as a function of samples $n$.
}
\else
The above discussion suggests that as a single DNN is trained,
it exhibits benign fitting early in training (when it has not fit the noise in its train set), and then transitions to tempered overfitting late in training (as it eventually fits the noise).
In Appendix \ref{app:time_dynamics}, we show a simple experiment in which an MLP trained on synthetic noisy data clearly exhibits this transition between regimes.
\fi

\ifarxiv
\paragraph{Remark on Shape of Profiles.}
The fact that noise profiles for classification all seem to approach lines with slope one is
is not at all trivial --- many alternate profile shapes are a priori possible.
This empirical observation was formalized in \citet{nakkiran2020distributional},
but remains theoretically open to understand.
In the KR setting, Theorem \ref{thm:trichotomy}b states that test error will
indeed be proportional to target noise, but the proportionality constant will generally be smaller than one.
We suggest this difference in noise sensitivity is one important difference between classification and regression.
\fi

\section{Limitations}
\label{sec:limitations}

In this paper, we have presented a taxonomy of overfitting, presented empirical evidence that DNNs trained to interpolation exhibit tempered overfitting, and identified spectral conditions under which KR, a toy model for DNNs, falls into each regime.
This is a first study into this regime, and many questions await careful exploration.
We detail several here.

First, our DNN results are entirely empirical, and there is room for complementary theoretical studies into the overfitting regimes of shallow networks and wide networks with NTK and mean-field parameterization.
Second, while we have used several real and synthetic datasets, all are of at most moderate size, and a study at large scale --- using correspondingly large models --- is potentially interesting.
Third, we have found that in some realistic settings, interpolating DNNs are tempered, but it remains open whether there might exist settings or tasks for which interpolating DNNs overfit benignly or catastrophically.
Fourth, Theorem \ref{thm:trichotomy} required a “universality” assumption which ought to be checked for each kernel.
Finally, our KR results suggest that input or manifold dimension should play a role in the degree of tempered overfitting, and the effect of dimensionality on overfitting is yet to be disentangled.

\section{Conclusion}
\label{sec:conclusions}

In this paper we study the nature of overfitting in learning methods which interpolate their training data.
Much of contemporary theory and experimental work categorizes overfitting as either catastrophic or benign.
In contrast, we observe that many natural learning procedures,
including DNNs used in practice, overfit in a manner that is neither benign nor catastrophic--- but rather in an intermediate regime.
We identify and formally define this regime, which we call {\it tempered overfitting}. 
We present empirical evidence of learning procedures that exhibit tempered overfitting, on both synthetic and natural data, using kernel machines and deep neural networks.
We show tempered overfitting can be quantified in terms of noise profiles,
which measure how asymptotic performance depends on noise in the train distribution.
For kernel regression, we provide a theoretical result in the form of a trichotomy:
conditions on problem parameters which yield each of the three regimes of overfitting.

Our work presents an initial study of tempered overfitting
and lays the framework for future study of what we believe is a rich and relevant regime for
modern learning.
We hope this framework inspires further investigation into tempered overfitting
for more complex models, both theoretically and experimentally.
For example, it is open to understand which conditions on neural network architecture, training hyperparameters
and data distribution lead to benign, tempered, or catastrophic overfitting, in analogy to our ``kernel trichotomy'', and an answer might shed light on practical DNNs.

\subsubsection*{Acknowledgments}
The authors thank Nikhil Ghosh, Annabelle Carrell, Spencer Frei, Daniel Beaglehole, Gil Kur, and Russ Webb for useful discussion and feedback on the manuscript.
JS additionally thanks two strangers on a DC Metro for useful discussion.
PN especially thanks the Simons Institute, UCSD, and Apple MLR for collaborative environments which made this project possible.

We are grateful for support from the National Science Foundation (NSF) and the Simons Foundation for the Collaboration on the Theoretical Foundations of Deep Learning\footnote{\url{https://deepfoundations.ai/}} through awards DMS-2031883 and \#814639 as well
as NSF IIS-1815697 and the TILOS institute (NSF CCF-2112665).
NM is thankful to be funded and supported for this research by the Eric and Wendy Schmidt Center.
JS gratefully acknowledges support from the NSF Graduate Fellow Research Program (NSF-GRFP) under grant DGE 1752814.
This work used the Extreme Science and Engineering Discovery Environment (XSEDE) \citep{XSEDE}, which is supported by NSF grant number ACI-1548562, Expanse CPU/GPU compute nodes, and allocations TG-CIS210104 and TG-CIS220009.

\subsubsection*{Author Contributions}
NM developed and performed the bulk of the experiments, drafted the initial version of the manuscript, actively contributed to writing and editing, and led the team logistically.
JS proposed and proved the theoretical KR results, aided in experimentation,
and contributed to the writing throughout the paper.
AA developed and executed most of the kernel experiments, provided support to other experiments, and helped edit the manuscript.
PP and MB provided guidance for experiments and theory, and reviewed and edited drafts of the manuscript.
PN initiated this research agenda and provided active input throughout the experimentation, theory, and writing phases of the project.

\ifarxiv
\newpage
\fi

\bibliographystyle{plainnat}
\bibliography{refs}

\newpage
\appendix

\section{Proof of \texorpdfstring{\Cref{thm:trichotomy}}{}: KR Trichotomy}
\label{app:kr_proofs}

In this appendix, we provide proofs and further discussion of our theoretical results for KR. 
To aid readers, we begin by reproducing Equation \ref{eqn:eigenlearning} for the approximate generalization MSE of KR:

\begin{equation} \label{eqn:eigenlearning_repr} \tag{\ref{eqn:eigenlearning}}
    \begin{split}
        \e \approx \e_0 \left( \sum_i (1 - \mc L_i)^2 v_i^2 + \epsilon^2 \right),
        \ \ \ \ \text{where} \ \ \ \
        \e_0 \equiv \frac{n}{n - \sum_j \mc L_j^2}, \\
        \mc L_i \equiv \frac{\lambda_i}{\lambda_i + \C},
        \ \ \ \ \text{and} \ \ \ \
        \C \ge 0 \text{ satisfies } \sum_i \frac{\lambda_i}{\lambda_i + \C} + \frac{\delta}{\C} = n.
    \end{split}
\end{equation}

The works obtaining this result \citep{bordelon:2020-learning-curves,canatar:2021-spectral-bias,jacot:2020-KARE,simon:2021-eigenlearning} make certain approximations common in the statistical physical literature.
The main one is the \textit{universality} assumption that the eigenfunctions can be replaced by structureless Gaussian random variables without changing the statistics of test risk (see \citet{jacot:2020-KARE} for a formalization of this assumption).
Though this step discards most of the information in the problem (leaving only the scalar eigenvalues and eigencoefficients), it is validated by the close match between experiment and the theory derived with this approximation.
KR is equivalent to linear regression in the high-dimensional kernel embedding space defined by its eigenfeatures, and we note several recent works which have derived the same equations for high-dimensional linear regression, analogously assuming i.i.d. random covariates \citep{hastie2019surprises, bartlett2021deep}.
This assumption of i.i.d. random covariates is also made by \citet{bartlett2019benign}.

In studying these equations, we assume the following:

\begin{assumption} \label{assumption:vals_and_coeffs}
The kernel eigenvalues $\{\lambda_i\}_{i=1}^{\infty}$ and target function eigencoefficients $\{v_i\}_{i=1}^\infty$ satisfy
\begin{enumerate}[(a)]
    \item $\lambda_i > \lambda_j$ if $i > j$,
    \item $\sum\limits_{i=1}^\infty \lambda_i < \infty$,
    % \item $\lim_{i \rightarrow \infty} \sum\limits_{j \ge i} \lambda_i = 0$,
    \item $\{\lambda_i\}_{i=1}^{\infty}$ contains infinitely many nonzero elements,
    \item $\sum\limits_{i=1}^\infty v_i^2 < \infty$, and
    \item $\lim\limits_{\lambda \rightarrow 0^+} \sum\limits_{i | \lambda_i < \lambda} v_i^2 = 0$.
\end{enumerate}
\end{assumption}

These natural assumptions imply that
(a) we have indexed the eigenvalues in descending order,
(b) the trace of the kernel is finite,
% (c) the kernel does not place non-negligible trace in arbitrarily-low eigenvalues\footnote{Kernels violating Assumption \ref{assumption:vals_and_coeffs}c can be shown to generalize as if the nonzero asymptotic tailsum were a ridge parameter.},
(c) the rank of the kernel is infinite (which is typically the case in practice),
(d) the target function has finite $\ell^2$-norm, and
(e) the target function does not place non-negligible weight in arbitrarily-low (or zero) eigenmodes\footnote{Weight placed in zero eigenmodes (i.e. lying outside the RKHS of the kernel) should be regarded as noise and included in $\epsilon^2$ instead of $\{v_i\}_i$.}.
While condition (e) is somewhat nonstandard, we note that it is strictly \textit{weaker} than requiring finite RKHS norm of $f$ w.r.t. $K$:
\begin{equation}
    |\!| f |\!|_K^2 = \sum_{i=1}^\infty \frac{v_i^2}{\lambda_i} < \infty.
\end{equation}
We note that all eigenvalues are nonnegative because the kernel is positive semidefinite.

The constant $\C$ fixed by Equation \ref{eqn:eigenlearning_repr} satisfies

\begin{align}
    % C &\le \frac{1}{n} \left( \delta + \sum_{i \ge 1} \lambda_i \right), \label{eqn:C_ub_1} \\
    % C &\le \delta + \sum_{i \ge n} \lambda_i, \\
    \C &\le \frac{1}{n \gamma} \left( \delta + \sum_{i \ge n(1 - \gamma)} \lambda_i \right), \label{eqn:C_ub_2} \\
    \C &\ge \gamma \lambda_{n (1 + \gamma)}, \label{eqn:C_lb_1}
\end{align}

where $\gamma \in (0,1)$ in Equation \ref{eqn:C_ub_2}, $\gamma \in (0,\infty)$ in Equation \ref{eqn:C_lb_1}, and in both cases $\gamma$ satisfies $\gamma n \in \mathbb{Z}$. These bounds follow from more general bounds in \cite{simon:2021-eigenlearning}.
We will use them shortly.

\textbf{Proof of Theorem \ref{thm:trichotomy}}. We begin by observing from Equation \ref{eqn:C_ub_2} with any $\gamma \in (0,1)$ that $\C = \mathcal{O}(1/n)$, and thus $\lim_{n \rightarrow \infty} \C = 0$ and $\lim_{n \rightarrow \infty} \mc L_i = 1$ if $\lambda_i$ is bounded away from 0.
Paired with Assumption \ref{assumption:vals_and_coeffs}e, this implies that $\lim_{n \rightarrow \infty} \sum_i (1 - \mc L_i)^2 v_i^2 = 0$, and thus, examining Equation \ref{eqn:eigenlearning_repr}, we find that asymptotic MSE is dominated by the noise $\epsilon^2$ and given by
\begin{equation}
    \lim_{n \rightarrow \infty} \e = \left( \lim_{n \rightarrow \infty} \e_0 \right) \epsilon^2.
\end{equation}
Asymptotic MSE is thus dominated by the noise, and we can neglect the target coefficients $\{v_i\}_i$.

We now prove each clause of the theorem by finding $\lim\limits_{n \rightarrow \infty} \e_0$, beginning with the two parts of clause (a).

\textit{\textbf{Clause (a):} if $\delta > 0$, then $\lim\limits_{n\rightarrow\infty} \e = \epsilon^2$.}

We assume that $\delta > 0$.
To begin, we note a lower bound for $\C$.
Since
\begin{equation}
    n = \sum_{i=1}^\infty \frac{\lambda_i}{\lambda_i + \C} + \frac{\delta}{\C} \ge \frac{\delta}{\C},
\end{equation}
it follows that $\C \ge \delta / n$.
Using this bound, we find that
\begin{equation} \label{eqn:L2_sum_bound}
    \frac{1}{n} \sum_{i=1}^\infty \mc L_i^2
    = \frac{1}{n} \sum_{i=1}^\infty \frac{\lambda_i^2}{(\lambda_i + \C)^2}
    \le \gamma + \frac{1}{n} \sum_{i \ge n\gamma} \frac{\lambda_i^2}{\C^2}
    \le \gamma + n \sum_{i \ge n\gamma} \frac{\lambda_i^2}{\delta^2},
\end{equation}
where $\gamma \in (0,1)$ and we have ``given up" on the first $\gamma n$ terms in the sum, replacing them with 1.

Because $\sum_{i=1}^\infty \lambda_i < \infty$, it must hold that $\lambda_i = o(i^{-1})$, and thus the sum on the far RHS of Equation \ref{eqn:L2_sum_bound} approaches zero as $n \rightarrow \infty$.
This tells us that $\lim_{n \rightarrow \infty} \e_0 = \lim_{n \rightarrow \infty} n / (n - \sum_i \mc L_i^2) \le 1 / (1 - \gamma)$.
By choosing $\gamma$ to be arbitrarily small, we can push this bound to $1$, which proves the clause.

\textit{\textbf{Clause (a'):} if $\lambda_i = i^{-1} \log^{-\alpha} i$ for some $\alpha > 1$, then $\lim\limits_{n\rightarrow\infty} \e = \epsilon^2$.}

We first set $\gamma = 1$ in Equation \ref{eqn:C_lb_1} and find that $\C \ge \lambda_{2n} = (2n)^{-1} \log^{-\alpha}(2n)$.
Anticipating our next move, we then note that, if $i \ge n$,
\begin{equation} \label{eqn:logbound}
    \C i \log^\alpha i \ge \frac{\log^\alpha n}{2 (\log^\alpha n + \log^\alpha 2)} \ge \frac{1}{5} \ \ \ \ \ \text{for sufficiently large $n$}.
\end{equation}
We then observe that
\begin{equation}
    n = \sum_{i=1}^\infty \frac{1}{1 + \C i \log^\alpha i}
    \ge \sum_{i \ge n} \frac{1}{1 + \C i \log^\alpha i}
    \ge \frac{1}{6 \C} \sum_{i \ge n} \frac{1}{i \log^\alpha i}
    \ge \frac{1}{6 \C} \frac{1}{(\alpha - 1) \log^{\alpha - 1} n},
\end{equation}
where in the third step we have used Equation \ref{eqn:logbound} and in the fourth step we have used the fact that $\int_x^\infty z^{-1} \log^{-\alpha} z dz = (\alpha - 1)^{-1} \log^{-\alpha+1}x$.
This tells us that
\begin{equation} \label{eqn:logbound_2}
    \C \ge \left(6 (\alpha - 1) n \log^{\alpha - 1} n \right)^{-1}.
\end{equation}

We then observe that
\begin{equation} \label{eqn:lrn_sum_logbound}
    \sum_{i=1}^\infty \mc{L}_i^2
    \le \frac{n}{\log \log n}
    + \sum_{i \ge \frac{n}{\log \log n}} \frac{\lambda_i^2}{\C^2}
    = \frac{n}{\log \log n}
    + \C^{-2} \Theta\left( \frac{\log \log n}{n \log^{2\alpha} n} \right),
\end{equation}
where in the first step we have ``given up" on the first $\frac{n}{\log \log n}$ terms and in the second step we have used the fact that $\int_x^\infty z^{-2} \log^{-2 \alpha} z dz \xrightarrow[]{\text{large } x} x^{-1} \log^{-2 \alpha}x$.
Plugging in Equation \ref{eqn:logbound_2}, we find that the RHS of Equation \ref{eqn:lrn_sum_logbound} approaches $\frac{n}{\log \log n}$ at large $n$, and is thus $o(n)$.
This implies benignness as in the proof of clause (a).

\textit{\textbf{Clause (b):} if $\delta = 0$ and $\lambda_i = i^{-\alpha}$ for some $\alpha > 1$, then $\lim\limits_{n\rightarrow\infty} \e_0 = \alpha$.}

The desired limit follows from direct computation upon replacing the sum over eigenvalues in the definition of $\C$ with an integral.
To make the result rigorous, we shall simply bound this sum between two integrals.

The constant $\C$ satisfies
\begin{equation}
    \sum_{i = 1}^{\infty} \frac{i^{-\alpha}}{i^{-\alpha} + \C}
    = \sum_{i \ge 1}^{\infty} \frac{1}{1 + \C i^\alpha}
    = n.
\end{equation}

Noting that the summand decreases monotonically with $i$, we find that
\begin{equation}
    n = \sum_{i = 1}^{\infty} \frac{1}{1 + \C i^\alpha}
    \ge \int_1^\infty \frac{d i}{1 + \C i^\alpha}
    \ge \int_0^\infty \frac{d i}{1 + \C i^\alpha} - 1
\end{equation}

and

\begin{equation}
    n = \sum_{i = 1}^{\infty} \frac{1}{1 + \C i^\alpha}
    \le \int_0^\infty \frac{d i}{1 + \C i^\alpha}.
\end{equation}

It follows that $\C^{}_- \le \C \le \C^{}_+$, where $\C^{}_-$ and $\C^{}_+$ satisfy
\begin{align}
    \int_0^\infty \frac{d i}{1 + \C^{}_- i^\alpha} &= n+1, \label{eqn:C_b_-_def}\\
    \int_0^\infty \frac{d i}{1 + \C^{}_+ i^\alpha} &= n. \label{eqn:C_b_+_def}
\end{align}

Using the fact that $\int_0^\infty (1 + x^{\alpha})^{-1} dx = \alpha^{-1} \pi \csc(\pi / \alpha)$ when $\alpha > 1$, we find that

\begin{align}
    \C^{}_- = \left[ \frac{\pi}{\alpha} \csc(\pi/\alpha) \right]^{\alpha} n^{-\alpha}, \\
    \C^{}_+ = \left[ \frac{\pi}{\alpha} \csc(\pi/\alpha) \right]^{\alpha} (n+1)^{-\alpha}.
\end{align}

These bounds converge as $n \rightarrow \infty$, and thus asymptotically $\C \rightarrow \left[ \frac{\pi}{\alpha} \csc(\pi/\alpha) \right]^{\alpha} n^{-\alpha}$.

A similar argument using the integral
$\int_0^\infty (1 + x^{\alpha})^{-2} dx = \alpha^{-2} (\alpha - 1) \pi \csc(\pi / \alpha)$
yields that
\begin{equation}
    \sum_{i = 1}^\infty \mc L_i^2
    \approx \int_0^\infty \frac{d i}{(1 + \C i^\alpha)^2}
    \rightarrow \C^{-1/\alpha} \frac{\alpha - 1}{\alpha^2} \pi \csc(\pi / \alpha) \rightarrow n \frac{\alpha - 1}{\alpha},
\end{equation}
where in the last step we have inserted our asymptotic expression for $\C$.
We now find that $\e_0 = \frac{n}{n - \sum_i \mc L_i^2} \rightarrow \alpha$.

\textit{\textbf{Clause (c):} if $\delta = 0$ and $\frac{\lambda_i}{\lambda_{i+1}} \ge \frac{i^{-\log i}}{(i+1)^{-\log (i+1)}}$ for all $i$, then $\lim\limits_{n\rightarrow\infty} \e_0 = \infty$.}

We begin by observing that
\begin{equation} \label{eqn:e0_inv_bound}
    \frac{1}{\e_0} = \frac{1}{n} \left( n - \sum_j \mc L_j \right) = \frac{1}{n} \sum_j \mc L_j (1 - \mc L_j) \le  \frac{1}{n} \sum_{i \le n} (1 - \mc L_i) +  \frac{1}{n} \sum_{i > n} \mc L_i.
\end{equation}
We shall show that both terms on the RHS of Equation \ref{eqn:e0_inv_bound} approach zero as $n \rightarrow \infty$, and thus $\e_0$ diverges.

Considering the first term first, we note that
\begin{align} \label{eqn:L_sum_1_bound}
    \frac{1}{n} \sum_{i \le n} (1 - \mc L_i)
    = \frac{1}{n} \sum_{i \le n} \frac{\C}{\lambda_i + \C}
    &\stackrel{\mathclap{\normalfont\tiny\mbox{(1)}}}{\le}
    \gamma + \frac{\C}{n} \!\!\! \sum_{i \le n(1 - \gamma)} \!\! \frac{1}{\lambda_i} \\
    &\le \gamma + \frac{\C}{\lambda_{n(1-\gamma)}}
    \stackrel{\mathclap{\normalfont\tiny\mbox{(2)}}}{\le}
    \gamma + \frac{1}{n \gamma} \sum_{i \ge n(1-\gamma)} \frac{i^{-\log i}}{(n(1-\gamma))^{-\log(n(1-\gamma))}},
\end{align}
where in step (1) we have ignored the last $n\gamma$ terms of the sum, and also used the fact that $\C/(\lambda_i + \C) \le \C / \lambda_i$, and in step (2) we have used Equation \ref{eqn:C_ub_2} to upper-bound $\C$, with $\gamma$ a constant parameter we will later take to be small.
Fixing $\gamma$ and taking $n \rightarrow \infty$, the final RHS of Equation \ref{eqn:L_sum_1_bound} approaches $\gamma$.
By making $\gamma$ arbitrarily small, we can make this upper bound approach zero.

Now examining the second term in Equation \ref{eqn:e0_inv_bound}, we observe that
\begin{multline} \label{eqn:L_sum_2_bound}
    \frac{1}{n} \sum_{i > n} \mc L_i
    = \frac{1}{n} \sum_{i > n} \frac{\lambda_i}{\lambda_i + \C}
    \stackrel{\mathclap{\normalfont\tiny\mbox{(1)}}}{\le}
    \gamma + \frac{1}{n \C} \sum_{i > n(1+\gamma)} \lambda_i \\
    \stackrel{\mathclap{\normalfont\tiny\mbox{(2)}}}{\le}
    \gamma + \frac{1}{n \gamma \lambda_{n(1 + \gamma)}} \sum_{i > n(1+\gamma)} \lambda_i
    \le \gamma + \frac{1}{n \gamma} \sum_{i \ge n(1+\gamma)} \frac{i^{-\log i}}{(n(1+\gamma))^{-\log(n(1+\gamma))}},
\end{multline}
where in step (1) we have again ``given up" on $n\gamma$ terms of the sum and in step (2) we have used Equation \ref{eqn:C_lb_1} to lower-bound $\C$.
Using the same argument as above, the final RHS of Equation \ref{eqn:L_sum_2_bound} approaches $\gamma$, which can be made arbitrarily small.

Combining subresults and looking again at Equation \ref{eqn:e0_inv_bound}, we find that $\lim_{n\rightarrow\infty} \e_0^{-1}$ can be given an arbitrarily small upper-bound, and thus $\lim_{n\rightarrow\infty} \e_0 = \infty$, which proves the clause.
\pushQED{\qed}\popQED

\textbf{Remarks on proofs and proof techniques}.
Clause (a) shows that fitting is benign if $\delta > 0$, but it is not hard to show that fitting is also benign if $\delta = 0$ and instead $\delta' \equiv \lim_{j\rightarrow\infty} \sum_{i \ge j} \lambda_i > 0$.
This odd tailsum $\delta'$ can be shown to act like an effective ridge parameter in Equation \ref{eqn:eigenlearning_repr}, but with $\delta=0$ the resulting kernel interpolates the data.
One way to add such an effective ridge to a kernel $K$ is to replace it with $K'(x_1,x_2) \equiv K(x_1,x_2) + \delta' \mathbbm{1}_{x_1 = x_2}$, which (assuming train and test sets are disjoint) simply adds a ridge parameter $\delta'$ to Equation \ref{eqn:krr_def} defining KR.
The fact that many small eigenvalues can act like a ridge parameter is proven by and used in the derivations of \citet{simon:2021-eigenlearning}.

When proving clause (c), we used a ratio condition to capture the notion of super-powerlaw decay.
We found that this ratio condition easier to work with than the weaker requirement that $\lambda_i = \mathcal{O}(i^{-\log i})$.
We note that there was nothing special in the choice of $i^{-\log i}$, and any slower super-powerlaw decay, like $i^{- \log \log i}$, would have also sufficed.

\section{Powerlaw and Laplace kernel experiments}
\label{app:powerlaws}

In Section \ref{sec:kernel} of the main text, we show via Theorem \ref{thm:trichotomy} that KR with a kernel with a powerlaw spectrum with exponent $\alpha$ overfits target noise by a factor $(\alpha - 1)$ as $n \rightarrow \infty$, and we conclude that Laplace kernels, which are known to have powerlaw spectral tails, ought to obey this rule\footnote{As a reminder, the asymptotic MSE was $\alpha \sigma^2$, while the Bayes-optimal value was $\sigma^2$, so the excess risk is $(\alpha - 1)\sigma^2$ and we say the noise is overfit by a factor of $(\alpha - 1)$.}.
Here we provide experimental evidence for both these claims.

We start with KR with a powerlaw spectrum.
No simple kernel + domain pair gives an exact powerlaw spectrum $\lambda_i = i^{-\alpha}$, so we perform a synthetic experiment with Gaussian random eigenfunctions, which is essentially linear regression with random features as in the setting of \citet{bartlett2019benign}.
Letting $n$ denote the number of train samples, $n'$ denote the number of test samples, and $M \gg n$ denote the number of eigenmodes (a.k.a. features), we first sample train and test feature matrices $\Phi \in \mathbb{R}^{M \times n}$ and $\Phi' \in \mathbb{R}^{M \times n'}$ with entries drawn i.i.d. from $\mathcal{N}(0,1)$.
We then construct the train-train and test-train kernels as $K_{\text{tr-tr}} = \Phi^T \Lambda \Phi$ and $K_{\text{te-tr}} = \Phi'^T \Lambda \Phi$, respectively, where $\Lambda \equiv \text{diag}(\lambda_1, \cdots, \lambda_M)$ with $\lambda_i = i^{-\alpha}$.

The train and test targets are given by $\mathcal{Y}_\text{tr} = \Phi^T \mathbf{v} + \eta_{\text{tr}}$ and $\mathcal{Y}_\text{te} = \Phi'^T \mathbf{v} + \eta_{\text{te}}$, respectively, where $\mathbf{v} \in \mathbb{R}^M$ is a vector of target eigencoefficients and the noise vectors are sampled $\eta_{\text{tr}} \sim \mathcal{N}(0, \sigma^2 \mathbf{I}_n)$ and $\eta_{\text{tr}} \sim \mathcal{N}(0, \sigma^2 \mathbf{I}_{n'})$.
The predicted test labels are computed via KR as $\hat{\mathcal{Y}}_\text{te} = K_{\text{te-tr}} K_{\text{tr-tr}}^{-1}$ and the MSE is then $\frac{1}{n'} | \mathcal{Y}_\text{te}  - \hat{\mathcal{Y}}_\text{te} |^2$.

We run experiments varying $n$ and using $n' = 3000$, $M = 10^4$, $\mathbf{v}_i \sim i^{-2}$ normalized so $\sum_{i=1}^M \mathbf{v}_i^2 = 10$, $\sigma^2 = 1$, and varying $\alpha$.
The results, shown in Figure \ref{fig:powerlaw_kr}, confirm that, as $n$ grows, MSE approaches $\alpha \sigma^2$ in accordance with Theorem \ref{thm:trichotomy}b.

\ifarxiv
\else
\begin{figure}[t]
    \centering
    \includegraphics[width=1.0\textwidth]{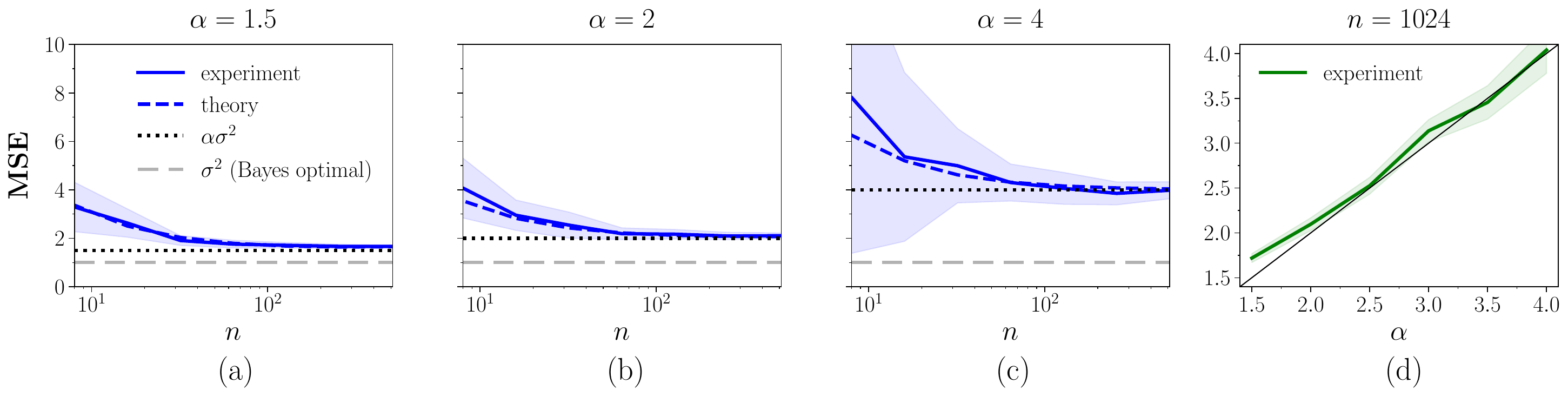}
    \caption{
    \textbf{As $n$ grows, the MSE of KR with Gaussian eigenfunctions and powerlaw kernel eigenspectra with exponent $\alpha$ approaches $\alpha \sigma^2$.}
    \textbf{(a-c)}: learning curves with different $\alpha$.
    \textbf{(d)}: test MSE at $n = 1024$ for varying $\alpha$, with the identity function shown by the solid line.}
    \label{fig:powerlaw_kr}
\end{figure}
\fi

\begin{figure}[t]
    \centering
    \includegraphics[width=.5\textwidth]{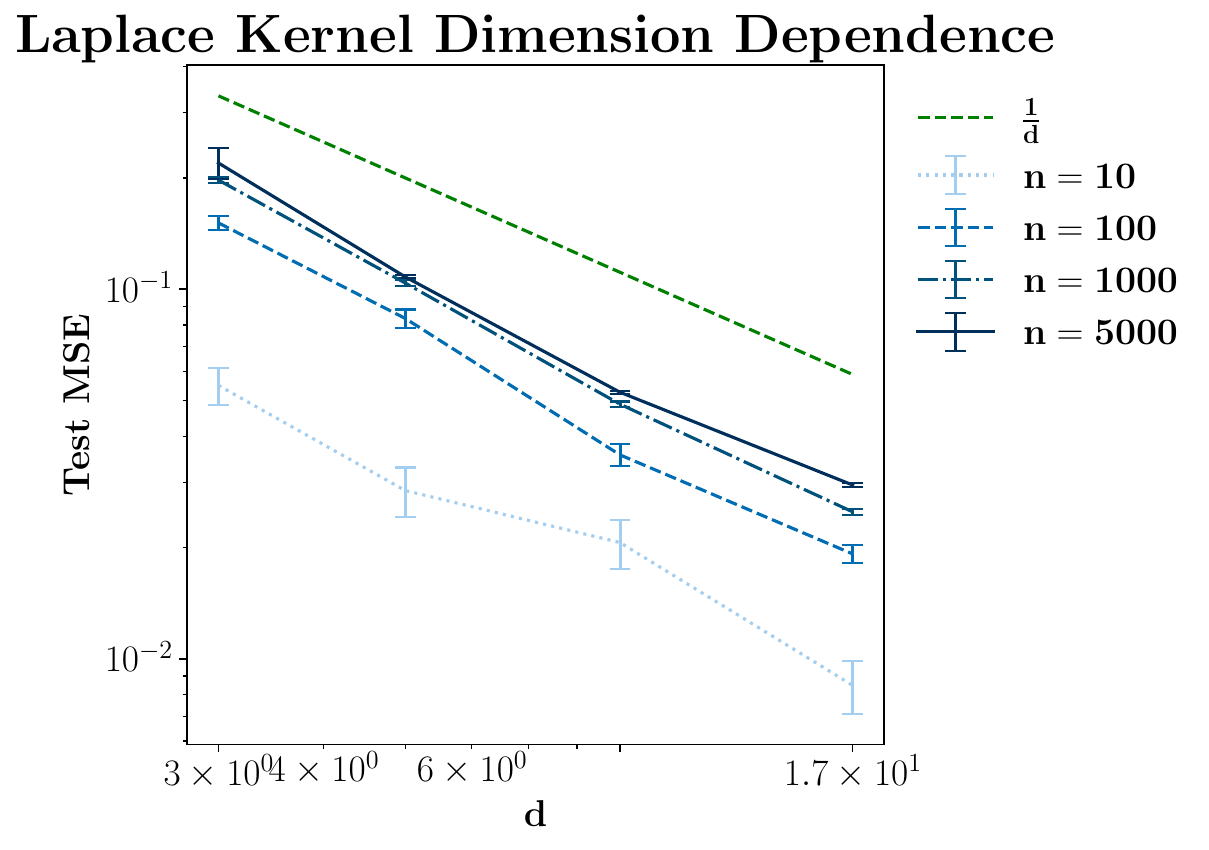}
    \caption{
    \textbf{As $n$ grows, the excess MSE of Laplace KR trained on noisy data on $\mathcal{S}^d$ decays like $1/d$.}
    }
    \label{fig:laplace_kr}
\end{figure}

We now study proper KR with a Laplace kernel and data sampled i.i.d. from the hypersphere $\mathcal{S}^d \equiv \{x \in \mathbb{R}^{d+1} | x^2 = 1\}$.
We run Laplace KR with varying $d$ and increasing $n$, using a kernel bandwidth of $1$, pure-noise train targets of $\mathcal{N}(0,1)$, and test samples with noiseless, uniformly-zero test targets so the resulting MSE is simply the excess risk.
As shown in Figure \ref{fig:laplace_kr}, we find that, for large $n$, excess MSE appears to decay as $\Theta(1/d)$ as expected from the kernel eigendecay\footnote{We note that it is unclear from this experiment whether excess MSE approaches $1/d$ \textit{exactly} as $n \rightarrow \infty$ or instead converges to roughly $c/d$ for some constant $c < 1$.}.

\section{Experimental Details}
\label{apdx:experiments}

\subsection{Datasets}
\label{apdx:datasets}

Our experiments are performed using the following real and synthetic datasets:
\begin{itemize}
    \item \textbf{The} $\mathbf{d}$\textbf{-sphere}: Data is sampled uniformly from $\mathcal{S}^{d}$ with noisy Gaussian targets.
    We use the convention that $d$ is the manifold dimension, and so $S^d$ is the hypersphere embedded in $\mathbb{R}^{d+1}$.
    \item \textbf{MNIST}. \citep{lecun1998mnist,lecun-mnisthandwrittendigit-2010}
    \item \textbf{Binary MNIST}. We binarize MNIST by classifying even vs. odd digits.
    \item \textbf{CIFAR-10}. \citep{krizhevsky2009learning}
    \item \textbf{Binary CIFAR-10}: We binarize CIFAR-10, forming two classes of animals (bird, cat, deer, dog) and vehicles (airplane, automobile, ship, truck).\footnote{We drop the ``frog" and ``horse" classes to create a balanced dataset.}
    \item \textbf{0/1 SVHN}. We binarize SVHN \citep{netzer2011reading}, selecting the digits zero and one. We use the ``train" and ``extra" datasets for train data from the \texttt{torchvision} library.
    \item \textbf{CINIC-10}. \citep{Darlow2018CINIC10IN} 10-class image dataset with samples sourced from CIFAR-10 and Tiny ImageNet.
\end{itemize}

We obtained the MNIST, CIFAR-10, and SVHN datasets from the \texttt{torchvision} library.\footnote{\url{https://pytorch.org/vision/stable/datasets.html}}

In all experiments with subsampled train datasets, we uniformly sample over all training data available, and experimental results are averaged over multiple samples (and random seeds).
All results are computed across the entire test set.

\subsection{Experimental Setups}
\label{apdx:experiment_setups}
Unless otherwise stated, models are trained with no weight decay, dropout, or other regularization techniques, as is done in \citet{zhang2016understanding}.
This encourages interpolation of training datasets even at high label noise levels.
As is standard, we perform channel-wise normalization of image datasets.

During training of Binary CIFAR-10, we additionally employ random horizontal flips. 
On MNIST, CIFAR-10 and SVHN, we do not use flips or crops.
As a result of this, and of our avoidance of regularization techniques, our models underperform current state-of-the-art methods, as is to be expected when intending to improve the state of understanding, not the state of the art.

We train the following models:

\textbf{Kernel Regression}:
For synthetic data on $\mathcal{S}^{d}$ we directly solve the kernel regression problem by inverting the data kernel matrix.
On MNIST, we create flattened vectors out of image data and train Gaussian and Laplacian kernels with stochastic gradient descent (SGD) in function space using EigenPro \citep{ma2017diving}. EigenPro automatically computes an optimal batch size given the number of training data points and available memory on the GPU being used.\footnote{\url{https://github.com/EigenPro/EigenPro-pytorch}}

\textbf{Neural Networks}:  We train multi-layer perceptrons (MLPs) and Wide ResNets (WRNs), and VGG networks \citep{simonyan2014very} using SGD with batch size 128, initial learning rate of 0.1, and momentum of 0.9. 
For the MLPs on synthetic data and WRNs on SVHN we employ a learning rate decay factor of 0.1; for WRNs on CIFAR-10 we use a decay factor of 0.2. 
Per-experiment details on specific learning rate schedules are given in Appendix \ref{apdx:experimental_details}.

Initial parameters for synthetic experiments are chosen by experimentation and largely guided by commonly used settings from non-synthetic cases.
For WRNs, the learning rate decay factors, momentum, and batch size choices for CIFAR-10 and SVHN are chosen based on \citet{zagoruyko2016wide}.
Wide ResNet model code is sourced from a publicly available Git repository.\footnote{\url{https://github.com/meliketoy/wide-resnet.pytorch/blob/master/networks/wide_resnet.py}}
MLP model code and training/testing scripts are written in-house. Neural networks and EigenPro kernels are trained using PyTorch\footnote{\url{https://pytorch.org/}}.

\subsection{Experiment-Specific Details}
\label{apdx:experimental_details}

\paragraph{MLP and Nearest Neighbor Noise Profiles on Binary MNIST (Figure \ref{fig:binary_mnist_mlp_noise_profile})}
We train $3 \times 1024$ MLPs, 1-NN, and $k$-NN ($k \sim \log n$) models on Binary MNIST. Results are averaged over five independent runs with mean and standard error bars reported.
Nearest Neighbor models use default settings from the \texttt{scikit-learn} library.\footnote{\url{https://scikit-learn.org/stable/}}
MLPs are trained using Adam, a learning rate of $1 \times 10^{-3}$, momentum of 0.9, batch size of 256. The learning rate is decayed by a factor of 0.1 at epochs 60 and 90. No learning rate warmup is used. All models are trained until the train loss reaches $1 \times 10^{-4}$ and achieve 100\% training accuracy.

\paragraph{Asymptotics of Kernel Regression on $d$-Sphere (Figure \ref{fig:kernel_trichotomy_plots})}
In this experiment, provided in Figure \ref{fig:kernel_trichotomy_plots}, we sample data, $\{x_i\}_i$, uniformly on $\mathcal{S}^d$ and train with pure noise target labels $y_i \sim \mathcal{N}(0, 1)$. 
Test MSE is computed on clean labels, $y_i = 0$.
In experiments with a ridge, we use $\delta=0.1$.
The kernel regression problem is solved by directly inverting the training data-data kernel matrix against the noisy labels.
For this plot the lines represent the median of 100 independent runs and error regions are given for 25\% and 75\% quantiles.

\begin{figure}[h]  
\centering
\begin{subfigure}{.45\textwidth}
  \centering
  \includegraphics[width=0.8\linewidth]{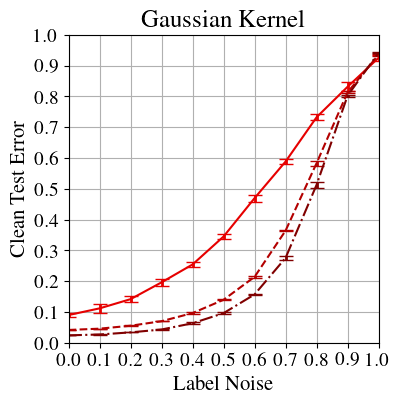}
  \caption{Gaussian Kernel}
  \label{fig:10mnist_gaussian}  
\end{subfigure}%
\begin{subfigure}{0.45\textwidth}
  \centering
  \includegraphics[width=0.8\linewidth]{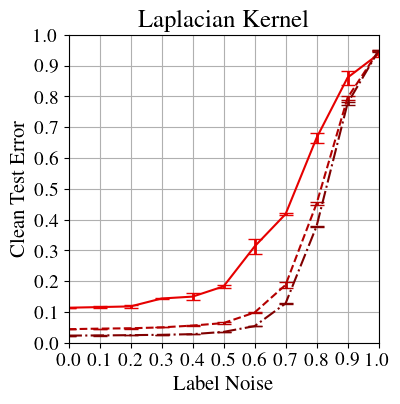}
  \caption{Laplacian Kernel}
  \label{fig:10mnist_laplace}%
\end{subfigure}
\begin{subfigure}{0.45\textwidth}
  \includegraphics[width=1.0\linewidth]{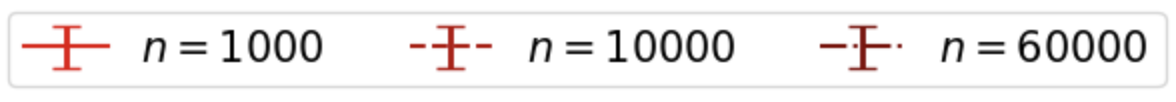}
  \label{fig:10mnist_legend}%
\end{subfigure}
\caption{We show noise profiles Gaussian and Laplacian kernels trained with label noise on 10-class MNIST for classification. We plot test classification error on the clean test set (no label noise) vs. training label noise. Overfitting is considered when the train MSE is $\leq 10^{-4}$. All models achieve 0 classification error on the training set.}
\label{fig:10mnist_noiseprofile}
\end{figure}

\paragraph{Kernel Noise Profiles on MNIST (Figure \ref{fig:10mnist_noiseprofile})}

Noise profiles for ridgeless Laplacian and Guassian kernels are provided for 10-class classification on MNIST.
All models reach 100\% training accuracy and $\leq 10^{-4}$ train MSE. Results are averaged over five independent runs.

\paragraph{MLP Noise Profiles on 10-Sphere (Figure \ref{fig:synthetic_noise_profile})} 
In this experiment, we take $f^*(x) = 1$ and inject label noise in the form of randomly flipping a fixed proportion, $p \in [0, 0.5]$, of training labels to $-1$.
The Bayes optimal classifier for $p < 0.5$ is $\mu(x) = 1$.
The goal of this experiment is to show that even in the simplest case of learning a constant function, interpolating neural networks suffer from noisy training data.
We present noise profiles for classification error as a function of training label flip probability in Figure \ref{fig:synthetic_noise_profile}.

The MLP is a $3 \times 1024$ network trained using SGD with initial learning rate of 0.1, momentum of 0.9, and batch size of 128. For $n<120000$ we cut the learning rate by a factor of 0.1 at epoch 150 and again at epoch 350. 
For $n=120000,360000$ we cut the learning rate by a factor of 0.1 at epoch 500 and again at epoch 750.

In all experiments, the stopping criteria is a train MSE loss $\leq 1e^{-4}$, regardless of where the learning rate schedule has reached by that point. 
Controlling for train loss stopping point allows us to meaningfully compare models that have achieved the ``same level of overfitting" and not introduce confounding factors due to late stage training effects. We average results over five independent runs.

\paragraph{WRN Noise Profiles on (Binary) CIFAR-10 and SVHN (Figures \ref{fig:intro_profile}, \ref{fig:profile_wideresnet_cifar10}, \ref{fig:svhn_profile})}

In all experiments, we train the Wide ResNet $28 \times 10$ for 60 total epochs using SGD with initial learning rate of 0.1, momentum of 0.9, and batch size of 128. 
We cut the learning rate by a factor of 0.2 in CIFAR-10 experiments and 0.1 in SVHN experiments. 
The learning rate in both settings is cut once at epoch 30, and again at epoch 40.

Classification label noise is injected for each point $\{(x_i, y_i)\}_i$ by uniformly re-sampling labels from alternative class labels, excluding the ground truth label. We resample labels in this way for a fraction of the dataset, $p$. For 10 class and binary classification we vary $p \in [0.0, 0.9]$ and $p \in [0.0, 0.5]$, respectively. For example, $p=0.9$ indicates that we train with exactly $90\%$ label noise.

Test classification error is computed on the clean test set. 
For Binary CIFAR-10, CIFAR-10, and 0/1 SVHN we average results over three independent runs and report mean and standard error bars.
On 0/1 SVHN, for each noise level and train set size, $n$, we perform 2-fold cross-validation to select the early-stop epoch which maximizes validation classification error averaged over both folds.
We additionally do not consider the first ten epochs for early-stopping, during the learning rate warmup phase.
We create the early-stopped noise profile from the classification test error attained at each of the early-stopped epoch choices.

\paragraph{MLP Error vs. Time (Figure \ref{fig:time_dependence})}

We train a $4 \times 512$ ReLU MLP trainsets of various sizes with $\{x_i\}$, sampled uniformly on $\mathcal{S}^4$.
Training labels are given by $y_i = 1 + \mathcal{N}(0, 1)$, a distribution for which the Bayes-optimal MSE is 1.
To reduce statistical error, we compute test MSE with clean labels $y_i = 1$ and simply add $1$ by hand to account for the noise.
We average over five trials and plot against $t \equiv \text{[learning rate]}\times\text{[epoch number]}$.
We train with full batch gradient descent using JAX and the neural-tangents library.\footnote{\url{https://github.com/google/neural-tangents}}

\paragraph{VGG Noise Profile on CINIC-10 (Figure \ref{fig:vgg_cinic10})}

We train a standard VGG-19 model with batch norm on the CINIC-10 dataset. We use SGD with initial learning rate 0.1 and momentum 0.9. We train for 200 epochs and decay the learning rate by a factor of 0.1 at epoch 60 and epoch 120.

\subsection{Compute Details}

Each neural network experiment and kernel experiment using EigenPro was run on a single NVIDIA V100 GPU. Experiments were performed in parallel using Expanse GPU nodes on XSEDE. \citep{XSEDE}
The final experiments for this paper took roughly 250 GPU-hours, and the experimentation phase of the project took roughly 1500 GPU-hours.

\section{Additional Experimental Results}

\subsection{Kernel Noise Profiles on MNIST}

We provide additional experimental results here for kernel noise profiles trained on MNIST data, as described in Appendix \ref{apdx:experimental_details}.

In Figure \ref{fig:10mnist_noiseprofile} we show noise profiles for Gaussian and Laplacian kernels trained on MNIST data at varying values of training data size, $n$. We immediately see from the plots that the Laplacian kernel is more tempered than the Gaussian kernel.

Note that while the Gaussian kernel does not look catastrophic, the input dimension of MNIST is larger than all experiments on $\cS^{d-1}$ shown in Figure \ref{fig:kernel_trichotomy_plots}.
In this case, we would need a significantly larger number of training samples to observe the Gaussian kernel entering the catastrophic regime.
Synthetic experiments on $\cS^{1000}$ (not shown) up to 1 million training samples show that the Gaussian kernel still does not enter the catastrophic regime, therefore with MNIST only containing 60k samples we see a more tempered noise profile.

\subsection{VGG Noise Profile}

\begin{figure}
    \centering
    \includegraphics[width=0.5\textwidth]{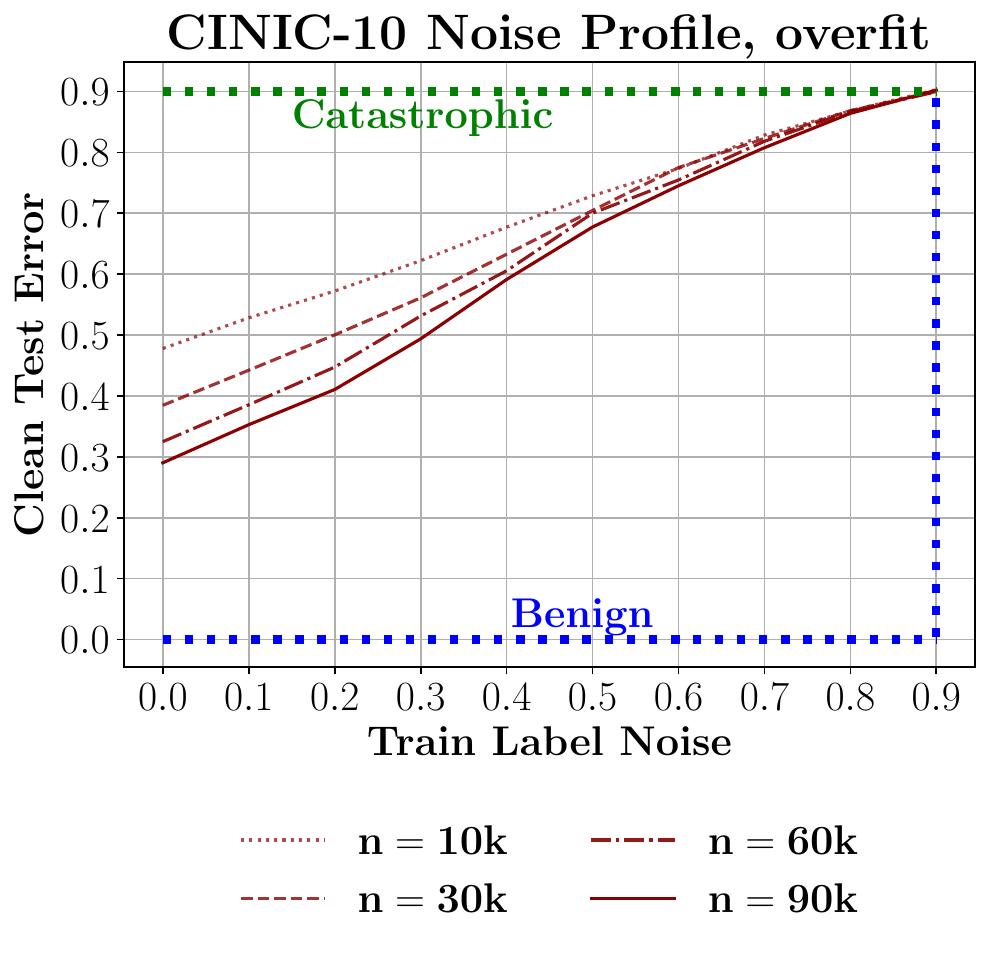}
    \caption{VGG-19 with batch norm overfit to the CINIC-10 dataset. We plot test classification error on the clean test set vs. training set label flip probability.}
    \label{fig:vgg_cinic10}
\end{figure}

We additionally explore noise profiles for overfit VGG networks on the CINIC-10 dataset. We do so in order to show that both residual and non-residual convolutional networks, on differing image datasets, maintain a tempered noise profile. In Figure \ref{fig:vgg_cinic10} we plot a 19-layer VGG network with batch norm overfit to the ten class CINIC-10 dataset.

On this dataset we were only able to run one experiment due to computational limitations, however in other experiments with deep neural networks we note low variance over the test error of DNNs at interpolation. Therefore, we expect the same to hold true over multiple runs of the VGG network on CINIC-10.

\end{document}